\documentclass[lettersize,journal]{IEEEtran}
\usepackage{amsmath,amsfonts}
\usepackage{array}
\usepackage[caption=false,font=normalsize,labelfont=sf,textfont=sf]{subfig}
\usepackage{textcomp}
\usepackage{stfloats}
\usepackage{url}
\usepackage{verbatim}
\usepackage{graphicx}
\usepackage{cite}
\usepackage{times}
\usepackage{epsfig}
\usepackage{graphicx}
\usepackage{amsmath}
\usepackage{amssymb}
\usepackage{booktabs}
\usepackage{multirow}
\usepackage{ulem}
\usepackage{color}
\usepackage{pifont}
\usepackage{balance}
\usepackage{colortbl}
\usepackage{xcolor}
\usepackage{makecell}
\usepackage{amsmath,amssymb,mathrsfs}
\usepackage{bm}
\usepackage{algorithm, algorithmic}
\newcommand{\cmark}{\ding{51}}

\usepackage{balance}
\usepackage{bbding} 
\usepackage{multirow}
\definecolor{newcolor}{rgb}{.8,.349,.1}

\begin{document}

\title{NDDepth: Normal-Distance Assisted Monocular Depth Estimation and Completion}

\author{Shuwei Shao, Zhongcai Pei, Weihai Chen$^{*}$, Peter C. Y. Chen and Zhengguo Li, \textit{Fellow, IEEE}
\thanks{This work was supported in part by the National Natural Science Foundation of China under grant 61620106012 and in part by the A*STAR Singapore under MTC Programmatic Funds grant M23L7b0021. }
\thanks{Shuwei Shao, Zhongcai Pei, Weihai Chen are with the School of Automation Science and Electrical Engineering, Beihang University, Beijing, China. (email: swshao@buaa.edu.cn, peizc@buaa.edu.cn, whchen@buaa.edu.cn)}
\thanks{Peter C. Y. Chen is with the Department of Mechanical Engineering, National University of Singapore, Singapore. (e-mail: mpechenp@nus.edu.sg)}
\thanks{Zhengguo Li is with the SRO department, Institute for Infocomm Research, A*STAR, Singapore. (ezgli@i2r.a-star.edu.sg)}}

\markboth{XXX, VOL. XX, NO. XX, XXXX 2023}
{Shao \MakeLowercase{\textit{et al.}}: NDDepth: Normal-Distance Assisted Monocular Depth Estimation and Completion}


\maketitle

\begin{abstract}
Over the past few years, monocular depth estimation and completion have been paid more and more attention from the computer vision community because of their widespread applications. In this paper, we introduce novel physics (geometry)-driven deep learning frameworks for these two tasks by assuming that 3D scenes are constituted with piece-wise planes. Instead of directly estimating the depth map or completing the sparse depth map, we propose to estimate the surface normal and plane-to-origin distance maps or complete the sparse surface normal and distance maps as intermediate outputs. To this end, we develop a normal-distance head that outputs pixel-level surface normal and distance. Meanwhile, the surface normal and distance maps are regularized by a developed plane-aware consistency constraint, which are then transformed into depth maps. Furthermore, we integrate an additional depth head to strengthen the robustness of the proposed frameworks. Extensive experiments on the NYU-Depth-v2, KITTI and SUN RGB-D datasets demonstrate that our method exceeds in performance prior state-of-the-art monocular depth estimation and completion competitors. The source code will be available at \url{https://github.com/ShuweiShao/NDDepth}.
\end{abstract}

\begin{IEEEkeywords}
Monocular depth estimation, Depth completion, Surface normal, Plane-to-origin distance, Piece-wise planar constraint 
\end{IEEEkeywords}

\section{Introduction}
\normalem
Active depth sensing has made substantial strides in performance and proven its practicality across various applications, including robotics~\cite{tateno2017cnn,jia2022object}, scene understanding~\cite{hazirbas2016fusenet} and augmented reality~\cite{lee2011depth}. Although specialist hardware sensors,~\textit{e.g.}, Microsoft Kinect and LiDAR, is able to capture accurate depth range, they tend to have difficulties in collecting dense depth maps because of the sensor noise, transparent and reflective surfaces or the limited number of scanning lines. Therefore, monocular depth estimation and completion capable of generating dense depth maps are developed.

Monocular depth estimation aims to predict the depth map from a single RGB image. Considering that a 2D image can be projected from infinite number of 3D scenes, such a task is indeed ill-posed and inherently ambiguous. Hence, solving it reliably demonstrates a formidable challenge for traditional methods~\cite{michels2005high,nagai2002hmm} because of their inherent limitations, typically involving low-dimensional and sparse distances or known and fixed objects. Recently, much progress has been made in this field benefiting from the explosion of deep learning~\cite{eigen2014depth,fu2018deep,lee2019big,bhat2021adabins,Yuan_2022_CVPR}. Most efforts focus on designing increasingly complicated and powerful networks, which renders the depth estimation a difficult fitting problem without the help of additional guidance. 

The mainstream of depth completion is to leverage the RGB image as guidance to complete the sparse depth map. A typical approach is to utilize multiple branches for extracting features from the sparse depth map and its corresponding RGB image, respectively and later merge them at different scales~\cite{tang2020learning, wong2021unsupervised,zhao2021adaptive,yan2022rignet}. In pursuit of advancing the frontiers, spatial propagation networks (SPNs)~\cite{cheng2018depth,park2020non,lin2022dynamic} and residual depth learning~\cite{gu2021denselidar, liu2021fcfr, liu2021learning} have been incorporated into the frameworks. Recently, Rho~\textit{et al.}~\cite{rho2022guideformer} and Zhang~\textit{et al.}~\cite{zhang2023completionformer} made use of a pure Transformer or the combination of Transformer and convolutional neural networks (CNNs) to further boost completion performance.

We argue that real-world 3D scenes,~\textit{e.g.}, indoor scenarios, typically exhibit a high degree of regularity and proper scene priors should be incorporated into the framework to improve the nature of the solution. Planes are a common representation for modeling geometric prior knowledge of 3D scenes~\cite{bodis2014fast,chauve2010robust,liu2019planercnn}. Patil~\textit{et al.}~\cite{patil2022p3depth} recently introduced a piece-wise planarity prior and adopted the offset vector field to borrow information from co-planar pixels. However, there is no direct constraint imposed on the offset vector field to help it learn about planar regions in their method. In addition, the planarity prior tends to fail in the high-curvature regions,~\textit{e.g.}, bushes, trees and other clutter from the outdoor scenarios, inevitably deteriorating the depth accuracy.

In this paper, we propose novel physics (geometry)-driven deep learning frameworks for monocular depth estimation and completion by assuming that 3D scenes are constituted with piece-wise planes. To be specific, we parametrize the plane representation via surface normal and plane-to-origin distance (the distance from the corresponding plane to the origin, i.e., camera center in our case) and develop a normal-distance head to output pixel-level surface normal and distance. Furthermore, a plane-aware consistency constraint is devised to enforce the surface normal and distance maps to be piece-wise constant. In order to obtain planar regions, we adopt the Felzenszwalb segmentation algorithm~\cite{felzenszwalb2004efficient} for online plane detection utilizing the geometric dissimilarity calculated from surface normal and distance maps. The regularized surface normal and distance maps are then transformed into depth maps. We note that the acquired depth maps are prone to make severe errors in the high-curvature regions owing to the invalidity of the planarity assumption. To account for such failure cases, we integrate a second depth head that is designed in accordance with the regular paradigms. The depth uncertainty is modeled to fully exploit the strengths of these two heads. For depth estimation, the depth and uncertainty maps are input into a developed contrastive iterative refinement module for depth refinement in a complementary manner. In terms of depth completion, the depth maps from these two heads are first fused utilizing the uncertainty maps. Then, the fused depth map and uncertainty map are input into the dedicated non-local spatial propagation network~\cite{park2020non} for depth refinement. The whole pipelines for monocular depth estimation and completion are presented in Figs.~\ref{Fig2} and~\ref{Fig12}, respectively.

The NDDepth approach was first proposed in our previous ICCV 2023 conference paper (Oral presentation)~\cite{Shao_2023_ICCV}. In this full version, we further extend NDDepth to depth completion. Different from previous approaches that directly complete the sparse depth map, we propose to complete the sparse surface normal and distance maps as intermediate outputs to exploit the geometry of the real-word 3D scenes. Besides, we find that only using the intermediate normal-distance representation can already achieve an impressive performance improvement. It is worth mentioning that while the state-of-the-art performance on the NYU-Depth-v2~\cite{silberman2012indoor} has been nearly saturated
for quite some time, such as from NLSPN~\cite{park2020non} (0.092 RMSE) to CompletionFormer~\cite{zhang2023completionformer} (CFormer, 0.090 RMSE), our NDDepth is capable of improving the CFormer from 0.090 RMSE to 0.081 RMSE, establishing a new record.

\begin{figure*}[!htb]
	\centering
	\includegraphics[width=0.95\linewidth]{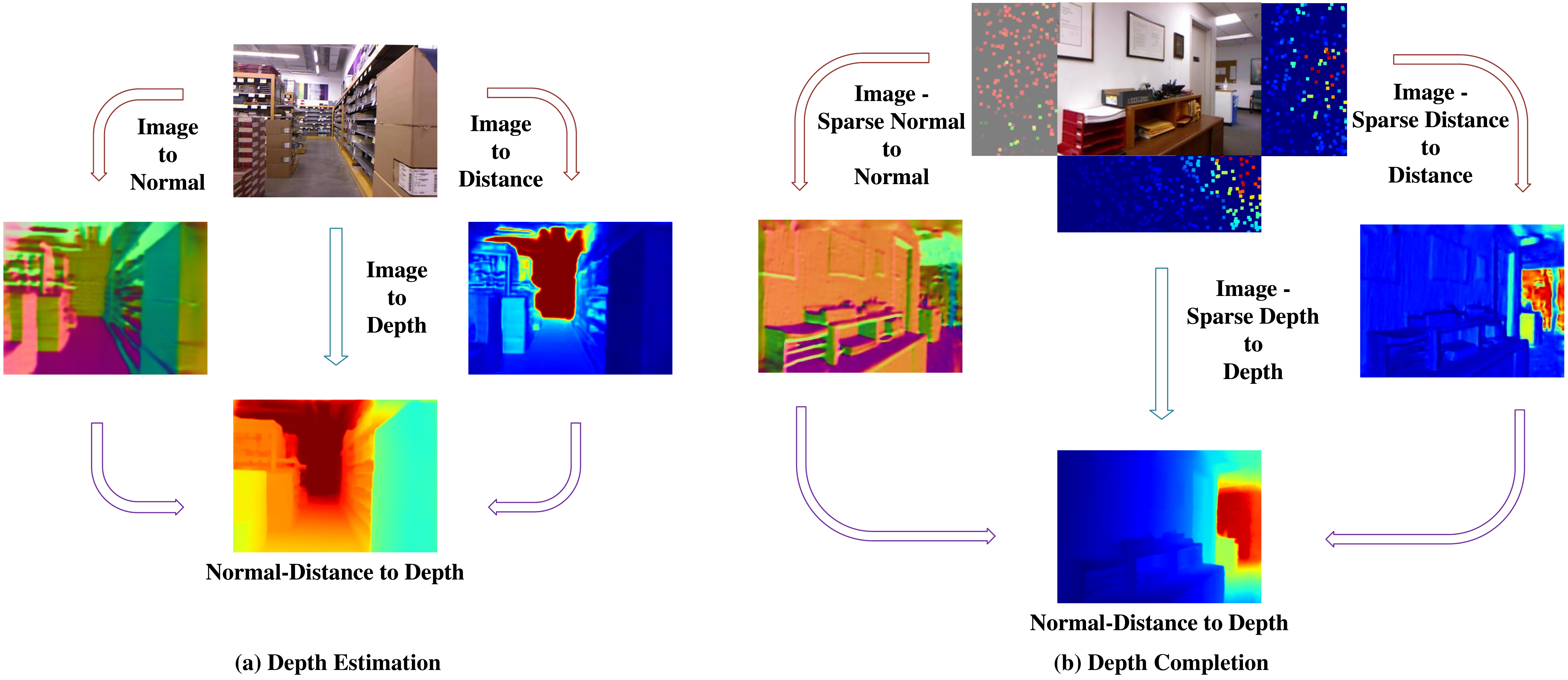}
	\caption{\textbf{A brief illustration of the relationship among RGB image, surface normal map/sparse surface normal map, plane-to-origin distance map/sparse plane-to-origin distance map and depth map/sparse depth map}. }
	\label{Fig1}
\end{figure*}

To summarize, the contributions of this work are listed as follows:
\begin{itemize}
	\item We introduce novel physics-driven deep learning frameworks for tasks of monocular depth estimation and completion, which consist of two heads built with asymmetric paradigms. The normal-distance head acquires piece-wise planar depth, while the regular depth head enhances the framework resilience.
	
	\item We develop a plane-aware consistency constraint to encourage the piece-wise constant property of surface normal and plane-to-origin distance maps, and a contrastive iterative refinement module to refine depth in a complementary fashion.
	
	\item Extensive experiments verify the efficacy of our designed components. The proposed method outperforms previous state-of-the-art monocular depth estimation and completion competitors on the NYU-Depth-v2, KITTI and SUN RGB-D datasets.
\end{itemize}

\section{Related work}
\subsection{Monocular Depth Estimation}
Monocular depth estimation involves the task of predicting the depth map from a single RGB image, which has witnessed dramatic progress over the years, with various methods being developed to tackle this challenge. In the early stages, Saxena~\textit{et al.}~\cite{saxena2005learning} considered both local and global image features and harnessed the power of a Markov Random Field to regress depth. Eigen~\textit{et al.}~\cite{eigen2014depth}, on the other hand, took a substantial leap forward by introducing CNN into depth estimation, using multi-scale networks to extract vital depth information. As the field evolved, Laina~\textit{et al.}~\cite{laina2016deeper} employed a fully convolutional network based on residual learning~\cite{He_2016_CVPR} and a reverse Huber loss for optimization. Later,  Cao~\textit{et al.}~\cite{cao2017estimating} and Fu~\textit{et al.}~\cite{fu2018deep} took a creative turn by reframing the depth regression problem as classification, making it more tractable. The classification-based depth estimation has inspired extensive follow-up works, for example, Adabins~\cite{bhat2021adabins} and BinsFormer~\cite{li2022binsformer}.~\cite{bhat2021adabins} observed that the depth distribution varies significantly between different images and proposed to generate adaptive bins in light of the image content.~\cite{li2022binsformer} further revisited~\cite{bhat2021adabins} by disentangling bins and probabilistic representations learning to prevent the global and fine-grained information from defacing each other. Besides, Agarwal~\textit{et al.}~\cite{agarwal2023attention} devised a bin center predictor by utilizing pixel queries at the coarsest level to predict adaptive bins. There are also a series of subsequent studies along the line of depth regression. Lee~\textit{et al.}~\cite{lee2019big} developed several multi-scale guidance layers to connect the intermediate layer features with the final layer depth prediction. Yang~\textit{et al.}~\cite{yang2021transformer} pioneered one of the endeavors by making use of the Vision Transformer (ViT)~\cite{dosovitskiy2020image} to capture the critical long-distance correlation in depth estimation. Yuan~\textit{et al.}~\cite{Yuan_2022_CVPR} selected the conventional path of Conditional Random Fields (CRFs) optimization and designed neural window fully-connected CRFs to reduce the computation complexity. However, the majority of these studies are data-driven approaches while the introduced estimation framework marries deep learning with the fundamental physics of the real-word 3D scenes.

\subsection{Depth Completion}

Image guided depth completion aims to predict the dense depth map from inputs of different modalities that include a sparse depth map and an RGB image. Following the advance of deep learning, depth completion has made significant strides recently. As one of the pioneering works, Ma~\textit{et al.}~\cite{ma2018sparse,ma2019self} adopted an encoder-decoder architecture to predict the dense output in either a supervised or self-supervised framework. To retain the precise measurements from the sparse depth input while further improving the final depth map quality, Cheng~\textit{et al.}~\cite{cheng2018depth} introduced the spatial propagation network (SPN)~\cite{liu2017learning} into depth completion. They developed a convolutional spatial propagation network (CSPN) and placed the CSPN behind the encoder-decoder network end to refine its prediction. Building upon CSPN, many follow-up works have sprung up. Cheng~\textit{et al.}~\cite{cheng2020cspn++} used learnable convolutional kernel sizes and number of iterations to improve the efficiency. Park~\textit{et al.}~\cite{park2020non} utilized deformable kernels~\cite{dai2017deformable} for propagation to alleviate the mixed-depth problem at object boundaries. Lin~\textit{et al.}~\cite{lin2022dynamic} leveraged independent affinity matrices in each propagation to relax the representation limitation. Zhang~\textit{et al.}~\cite{zhang2023completionformer} integrated Transformer and CNN to strengthen the capacity of the encoder-decoder network for better SPN refinement.
 
In addition to depending only on a single branch, the multi-branch networks have been employed to facilitate multi-modal fusion~\cite{zhang2018deep,van2019sparse,tang2020learning,hu2021penet,liu2021fcfr}. The conventional techniques for fusing multi-modal information involve feature concatenation or elementwise summation operations. Thereafter, more advanced fusion strategies have been introduced. Tang~\textit{et al.}~\cite{tang2020learning} applied the guided image filtering~\cite{he2012guided} while utilizing dynamically generated spatially-variant kernels. Zhong~\textit{et al.}~\cite{zhong2019deep} used channel-wise canonical correlation analysis. Zhang~\textit{et al.}~\cite{zhang2020multiscale} developed a neighbor attention mechanism to merge features at multiple scales. Zhao~\textit{et al.}~\cite{zhao2021adaptive} leveraged symmetric gated fusion in the decoder. More recently, Rho~\textit{et al.}~\cite{rho2022guideformer} designed separate Transformer branches to embed sparse depth map and RGB image, and a guided attention mechanism was proposed to capture inter-modal dependencies. Differently, we decouple depth completion into surface normal completion and plane-to-origin distance completion, to leverage the geometry of the real-world 3D scenes.

\begin{figure*}[!htb]
	\centering
	\includegraphics[width=0.95\linewidth]{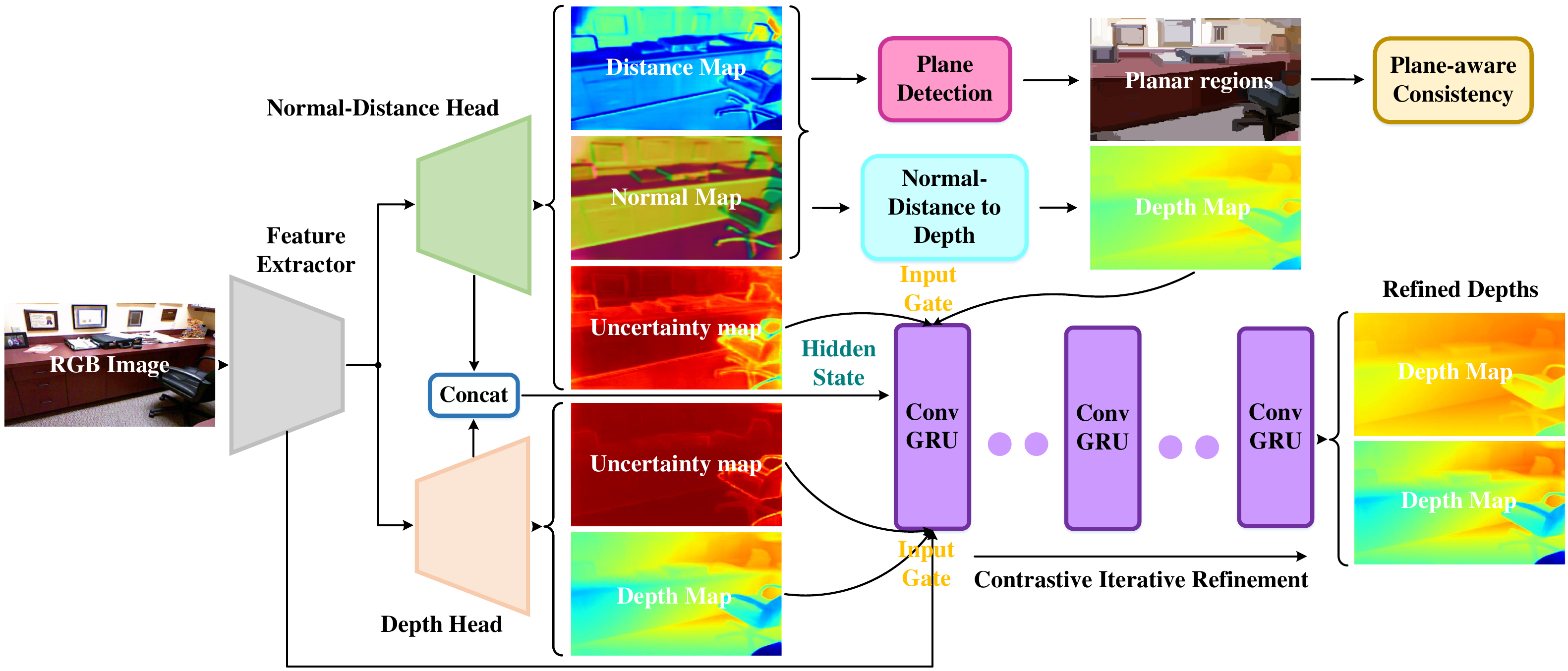}
	\caption{\textbf{Overview of the whole pipeline for monocular depth estimation}. We note that the contrastive iterative refinement occurs at a reduced 1/4 resolution and subsequently, the refined depth maps are upsampled to the full resolution using bilinear interpolation. For the sake of simplicity, we have omitted this particular detail in the flowchart.}
	\label{Fig2}
\end{figure*}

\subsection{Geometric Constraints for Depth}

Traditional methods, such as in multi-view stereo~\cite{gallup2010piecewise} and 3D reconstruction~\cite{chauve2010robust,bodis2014fast}, made use of the planarity prior to enable faster optimization and tackle poorly textured surfaces. Recently, Qi~\textit{et al.}~\cite{qi2018geonet} and Qiu~\textit{et al.}~\cite{qiu2019deeplidar} introduced surface normal to guide the depth prediction. Xu~\textit{et al.}~\cite{xu2019depth} leveraged the depth-normal constraints to refine coarse depth prediction in the plane-to-origin distance subspace. Kusupati~\textit{et al.}~\cite{kusupati2020normal} designed a consistency constraint between spatial depth gradients derived from depth and surface normal. Long~\textit{et al.}~\cite{Long_2021_ICCV} devised adaptive surface normal to determine the reliable local geometry. Huynh~\textit{et al.}~\cite{huynh2020guiding} and Yin~\textit{et al.}~\cite{yin2019enforcing} introduced non-local coplanarity constraints through either the depth-attention module or the virtual surface normal. Patil~\textit{et al.}~\cite{patil2022p3depth} proposed a piece-wise planarity prior by predicting the offset vector field to deliver information from co-planar pixels. Nevertheless, the offset vector field is not given any direct constraints to aid in its comprehension of planar regions. Moreover, the planarity prior is not always valid, leading to erroneous depth predictions in the high-curvature regions. By contrast, our method explicitly enforces the planar constraint inside each planar region. This effectively avoids the interference of pixels from other planes. Notably, the depth head in our framework allows such failure cases in~\cite{patil2022p3depth} to be mitigated.

\section{Methodology}
Real-world 3D scenes typically exhibit a significant degree of regularity, ~\textit{e.g.}, indoor scenarios. Therefore, it is reasonable to make the assumption that 3D scenes are constituted with piece-wise planes. In this section, we elaborate on the following parts, including depth from normal-distance constraint, plane-aware consistency, normal-distance assisted depth estimation and completion.

\subsection{Depth from Normal-Distance Constraint}
Let ${\bf{P}} = {\left[ {X,Y,Z} \right]^{\rm{T}}}$ denote a 3D point, and $\textbf{p} = {\left[ {u,v} \right]^{\rm{T}}}$ be its 2D projection on the image plane within a planar region of the 3D scenes. The surface normal is the vector originating from $\textbf{P}$ and perpendicular to the corresponding plane, denoted as $\textbf{N}\left( \textbf{p} \right)$. The distance from the origin (in our case, the camera center) to the corresponding plane is referred to as the plane-to-origin distance, denoted as $\mathcal{D}\left( \textbf{p} \right)$. Then, the normal-distance constraint is expressed as
\begin{equation} \textbf{N}\left( \textbf{p} \right)\textbf{P} = \mathcal{D}\left( \textbf{p} \right). \label{eq1} \end{equation}

In accordance with the fundamental principles of a pinhole camera, the mathematical representation of the projection from the 3D point $\textbf{P}$ to the 2D point $\textbf{p}$ is given by
\begin{equation} \textbf{D}\left( \textbf{p} \right)\widetilde {\textbf{p}} = {\textbf{K}}\textbf{P}, \label{eq2} \end{equation}
where $\textbf{D}\left( \textbf{p} \right)$ stands for the depth at $\textbf{p}$, $ \widetilde {\textbf{p}} $ denotes the homogeneous coordinate of $\textbf{p}$, and ${\textbf{K}}$ denotes the intrinsic matrix.

By further substituting Eq.~\ref{eq2} into Eq.~\ref{eq1}, the depth is derived via 
\begin{equation} \textbf{D}\left( \textbf{p} \right) = \frac{\mathcal{D}\left( \textbf{p} \right)}{{\textbf{N}\left( \textbf{p} \right){\textbf{K}^{ - 1}}\widetilde {\textbf{p}}}}. \label{eq4} \end{equation}

Rather than directly estimating the depth map or completing the sparse depth map,
we aim to estimate the surface normal and distance maps or complete the sparse surface normal and distance maps as intermediate outputs, and then apply Eq.~\ref{eq4} to derive the depth map. In contrast to the direct prediction of depth, the intermediate normal-distance representation enjoys the benefit of being piece-wise constant, which is not suitable for depth. This nature allows for the incorporation of a plane-aware consistency constraint that facilitates interactions among pixels, ultimately leading to an enhanced depth prediction.

\subsection{Plane-aware Consistency}
\textbf{Planar region detection}. To effectively enforce the plane-aware consistency constraint, it is imperative to identify planar regions. In line with previous works~\cite{concha2014using,concha2015dpptam,yu2020p,li2021structdepth}, we leverage the Felzenszwalb segmentation algorithm~\cite{felzenszwalb2004efficient} in our approach. 

Let \textbf{q} be an adjacent pixel to \textbf{p}. We define the dissimilarity in surface normal map between these two pixels as
\begin{equation}di{s_\textbf{N}}\left( {\textbf{p},\textbf{q}} \right) = \left\| {\textbf{N}(\textbf{q}) - \textbf{N}(\textbf{p})}\right\|,\end{equation}
where $\left\|  \cdot  \right\|$ denotes the Euclidean distance. Suppose $dis_\textbf{N}^{\max }$ and $dis_\textbf{N}^{\min }$ correspond to the maximum and minimum dissimilarities, respectively, among all adjacent pixels, which we use for normalizing the dissimilarity via
\begin{equation}{\overline {dis} _\textbf{N}} = \frac{{di{s_\textbf{N}}\left( {\textbf{p},\textbf{q}} \right) - dis_\textbf{N}^{\min }}}{{dis_\textbf{N}^{\max } - dis_\textbf{N}^{\min }}}.\end{equation}
We define the dissimilarity in plane-to-origin distance map as 
\begin{equation}di{s_\mathcal{D}}\left( {\textbf{p},\textbf{q}} \right) = \left| {{\mathcal{D}(\textbf{q}) - \mathcal{D}(\textbf{p})}} \right|. \end{equation}
Similarly, we normalize the distance dissimilarity via 
\begin{equation}{\overline {dis} _\mathcal{D}} = \frac{{di{s_\mathcal{D}}\left( {\textbf{p},\textbf{q}} \right) - dis_\mathcal{D}^{\min }}}{{dis_\mathcal{D}^{\max } - dis_\mathcal{D}^{\min }}}.\end{equation}

\begin{figure}[!htb]
	\centering
	\includegraphics[width=1.0\linewidth]{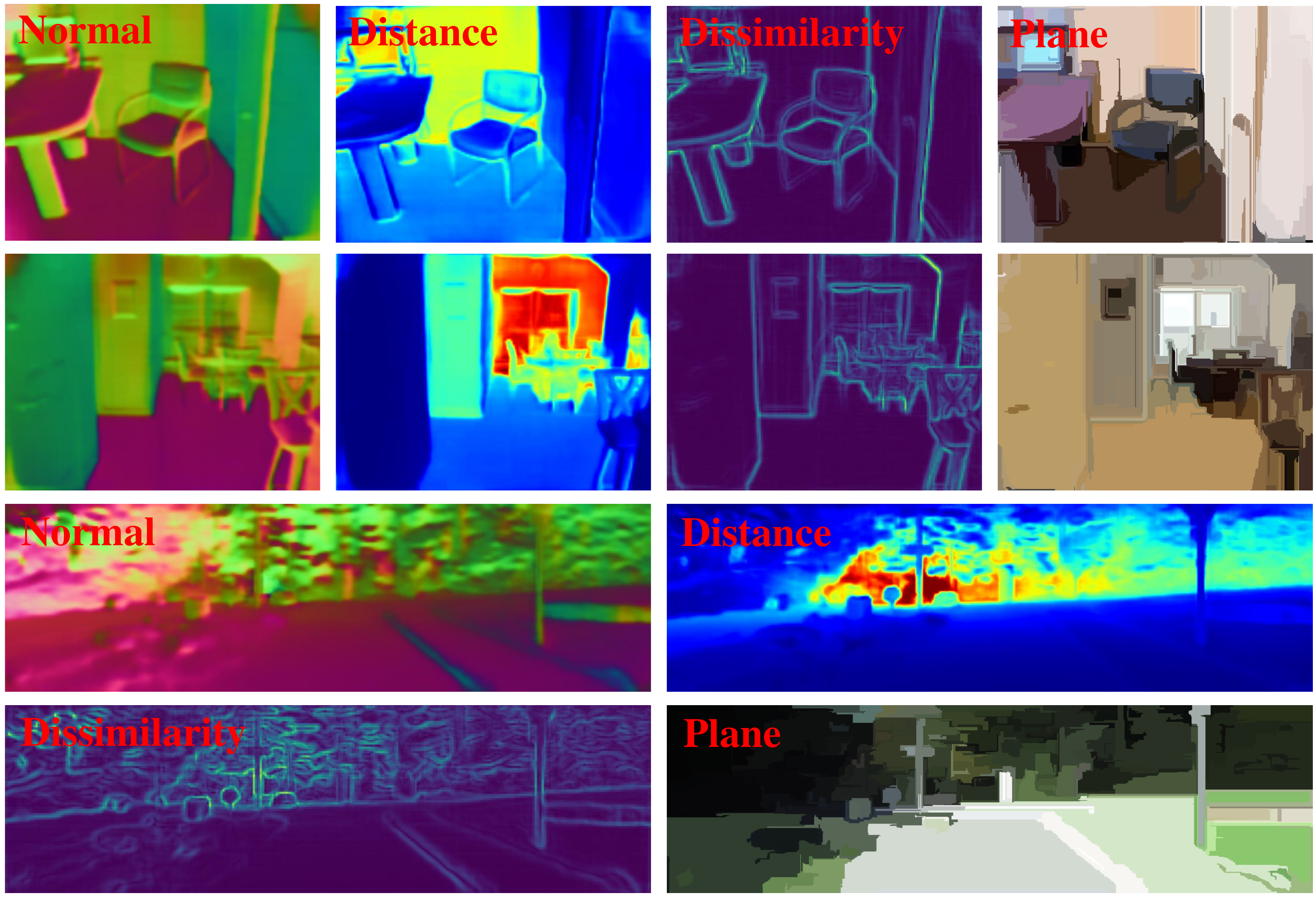}
	\caption{\textbf{Illustration of surface normal, plane-to-origin distance, geometric dissimilarity and detected planes}.  }
	\label{Fig4}
\end{figure}

The normalized surface normal and distance dissimilarities are combined to quantify the geometric dissimilarity,
\begin{equation}{\overline {dis} _g} = {\overline {dis} _\textbf{N}} + {\overline {dis} _\mathcal{D}}, \end{equation}

In light of the geometric dissimilarity, we adopt the Felzenszwalb segmentation algorithm to conduct online plane detection. Intuitively, planar regions,~\textit{e.g.}, indoor floors and walls, as well as outdoor roads, tend to occupy larger areas. Hence, we focus exclusively on regions exceeding a 200-pixel threshold for selection. In Fig.~\ref{Fig4}, we provide some qualitative results of surface normal, distance, geometric dissimilarity and detected planes. 

\textbf{Plane-aware consistency loss}. After identifying the planar regions, we encourage the plane-aware consistency by imposing penalties on the first-order gradients of surface normal and distance maps inside these planar regions, denoted as $\mathcal{M}\left( \textbf{p} \right)$,
\begin{equation} \resizebox{0.89\hsize}{!}{${\mathcal{L}_{pc}} = \sum\limits_\textbf{p} {\mathcal{M}\left( \textbf{p} \right){{ \left|\nabla\textbf{N}\left( \textbf{p} \right) \right|}}} + \sum\limits_\textbf{p} {\mathcal{M}\left( \textbf{p} \right) {{ \left|\nabla\mathcal{D}\left( \textbf{p} \right) \right|}} }$}. \end{equation}
The initial surface normal and distance predictions may not be accurate. Fortunately, they will gradually improve with each training epoch, resulting in a better detection, and vice versa.

\subsection{Normal-Distance Assisted Depth Estimation}

Given an RGB image $\textbf{R}$, previous methods train a network to directly map the RGB image into a depth prediction $\textbf{D}$, 
\begin{equation} {N_\theta }:\textbf{R} \to \textbf{D}. \end{equation} Different from previous studies, our approach uses a network that is equipped with a normal-distance head to predict surface normal and distance maps, \begin{equation} {N_\theta }: \textbf{R} \to \textbf{N}, \mathcal{D}, \end{equation}  
which are subject to regularization by the plane-aware consistency constraint. The regularized surface normal and distance maps are converted into depth predictions via Eq.~\ref{eq4}. However, the converted depth maps do not consistently produce accurate results and are susceptible to large errors in the high curvature regions. We observe that regular estimation paradigms tend to yield smaller errors in these regions. Therefore, we integrate a second depth head built upon~\cite{Yuan_2022_CVPR} into our framework. To fully capitalize on the strengths of both heads, we incorporate the depth uncertainty modeling. Then, a contrastive iterative refinement module is developed to enhance depth maps in a complementary fashion.

\textbf{Uncertainty modeling.} Identifying regions with large errors is critical before initiating the depth refinement. We model the depth uncertainty in the form of a probability density function of Laplace distribution~\cite{kendall2017uncertainties}, 
\begin{equation}
	\begin{array}{l}
	\textbf{U}_{1}^{gt}\left( \textbf{p} \right) = 1 - \exp \left( { - \frac{{\left| {{\textbf{D}_{1}}\left( \textbf{p} \right) - \textbf{D}^{gt}\left( \textbf{p} \right)} \right|}}{b }} \right)	\\
	\textbf{U}_{2}^{gt}\left( \textbf{p} \right) = 1 - \exp \left( { - \frac{{\left| {{\textbf{D}_{2}}\left( \textbf{p} \right) - \textbf{D}^{gt}\left( \textbf{p} \right)} \right|}}{b }} \right),
   \end{array}
\end{equation}
where $\textbf{D}_1 \left( \textbf{p} \right)$ and $\textbf{D}_2 \left( \textbf{p} \right)$ are the depth predictions of normal-distance and depth heads, respectively, $\textbf{D}^{gt}$ is the ground-truth depth map, $b$ stands for the error tolerance coefficient and is set to 0.2. As ${\textbf{D}^{gt}}\left( \textbf{p} \right)$ is not accessible at application phase, these two heads also predict uncertainty maps to approximate ${\textbf{U}_1^{gt}}$ and ${\textbf{U}_2^{gt}}$.  We denote the uncertainty predictions as $\textbf{U}_1\left( \textbf{p} \right)$ and $\textbf{U}_2\left( \textbf{p} \right)$, supervised by
\begin{equation}\resizebox{0.87\hsize}{!}{${\mathcal{L}_{\textbf{U}}} = \sum\limits_\textbf{p} {\left| {\textbf{U}_1\left( \textbf{p} \right) - {\textbf{U}_1^{gt}} \left( \textbf{p} \right)} \right|}  + \sum\limits_\textbf{p} {\left| {\textbf{U}_2\left( \textbf{p} \right) - {\textbf{U}_2^{gt}} \left( \textbf{p} \right)} \right|}$}.
\end{equation}

In addition to the uncertainty maps, we determine the complementary regions within these two depth maps by calculating the absolute difference between $\textbf{D}_1\left( \textbf{p} \right)$ and $\textbf{D}_2\left( \textbf{p} \right)$, 
\begin{equation}di{f_\textbf{D}}\left( \textbf{p} \right) = \left| {\textbf{D}_1 \left( \textbf{p} \right) - \textbf{D}_2 \left( \textbf{p} \right)} \right|, \end{equation}
where $di{f_\textbf{D}}$ is referred to as the complementary map.

\textbf{Contrastive iterative update.}
Grounded on the uncertainty map and complementary map, we refine depth maps iteratively. During each iteration, we update $\textbf{D}_1 \left( \textbf{p} \right)$ and $\textbf{D}_2 \left( \textbf{p} \right)$ as
\begin{equation} \resizebox{0.88\hsize}{!}{$\textbf{D}_1^{t + 1}\left( \textbf{p} \right) \leftarrow \textbf{D}_1^t\left( \textbf{p} \right) + \Delta \textbf{D}_1^t\left( \textbf{p} \right),\textbf{D}_2^{t + 1}\left( \textbf{p} \right) \leftarrow \textbf{D}_2^t\left( \textbf{p} \right) + \Delta \textbf{D}_2^t\left( \textbf{p} \right)$}, \end{equation}
where $\Delta \textbf{D}_1^t\left( \textbf{p} \right)$ and $\Delta \textbf{D}_2^t\left( \textbf{p} \right)$ are the depth updates at iteration $t$. Drawing inspiration from~\cite{teed2020raft}, we leverage a convolutional gated recurrent unit (ConvGRU)~\cite{chung2014empirical} to produce these updates. The ConvGRU is capable of preserving historical states and effectively exploiting the temporal context information during the refinement process. 

To begin, we employ two convolutional layers to transform $\textbf{D}_1^t$, $\textbf{D}_2^t$, $\textbf{U}_1$, $\textbf{U}_2$ and $dif_\textbf{D}$ into the feature space, respectively, and employ one convolutional layer to aggregate the projected features. Then, we concatenate the aggregated features with the image contextual feature from the feature extractor as input $\textbf{I}^t$. The structure inside ConvGRU is as follows,
\begin{equation}
	{{\bm{{\rm z}}}^{t+1}} = \sigma \left( {Con{v_{5 \times 5}}\left( {\left[ {{{\bm{{\rm h}}}^{t}}, \textbf{I}^t} \right], W_z} \right)} \right),
\end{equation}
\begin{equation}
	{{\bm{{\rm r}}}^{t+1}} = \sigma \left( {Con{v_{5 \times 5}}\left( {\left[ {{{\bm{{\rm h}}}^{t}}, \textbf{I}^t} \right]}, W_r\right)} \right),
\end{equation}
\begin{equation}
	{\widehat {\bm{{\rm h}}}^{t+1}} = \tanh \left( {Con{v_{5 \times 5}}\left( {\left[ {{{\bm{{\rm r}}}^{t+1}} \odot {{\bm{{\rm h}}}^{t}}, \textbf{I}^t}\right]}, W_h  \right)} \right),
\end{equation}
\begin{equation}
	{{\bm{{\rm h}}}^{t+1}} = \left( {1 - {{\bm{{\rm z}}}^{t+1}}} \right) \odot {{\bm{{\rm h}}}^{t}} + {{\bm{{\rm z}}}^{t+1}} \odot {\widehat {\bm{{\rm h}}}^{t+1}},
\end{equation}
where $Con{v_{5 \times 5}}$ is the separable ${5 \times 5}$ convolution, $\odot$ denotes the element-wise multiplication, $\sigma$ and tanh denote the sigmoid and tanh activation functions, respectively. The hidden state ${{\bm{{\rm h}}}^{t}}$ is initialized with the penultimate layer feature maps from both normal-distance and depth heads, concatenated together and activated with the tanh function.

\begin{figure*}[!htb]
	\centering
	\includegraphics[width=0.95\linewidth]{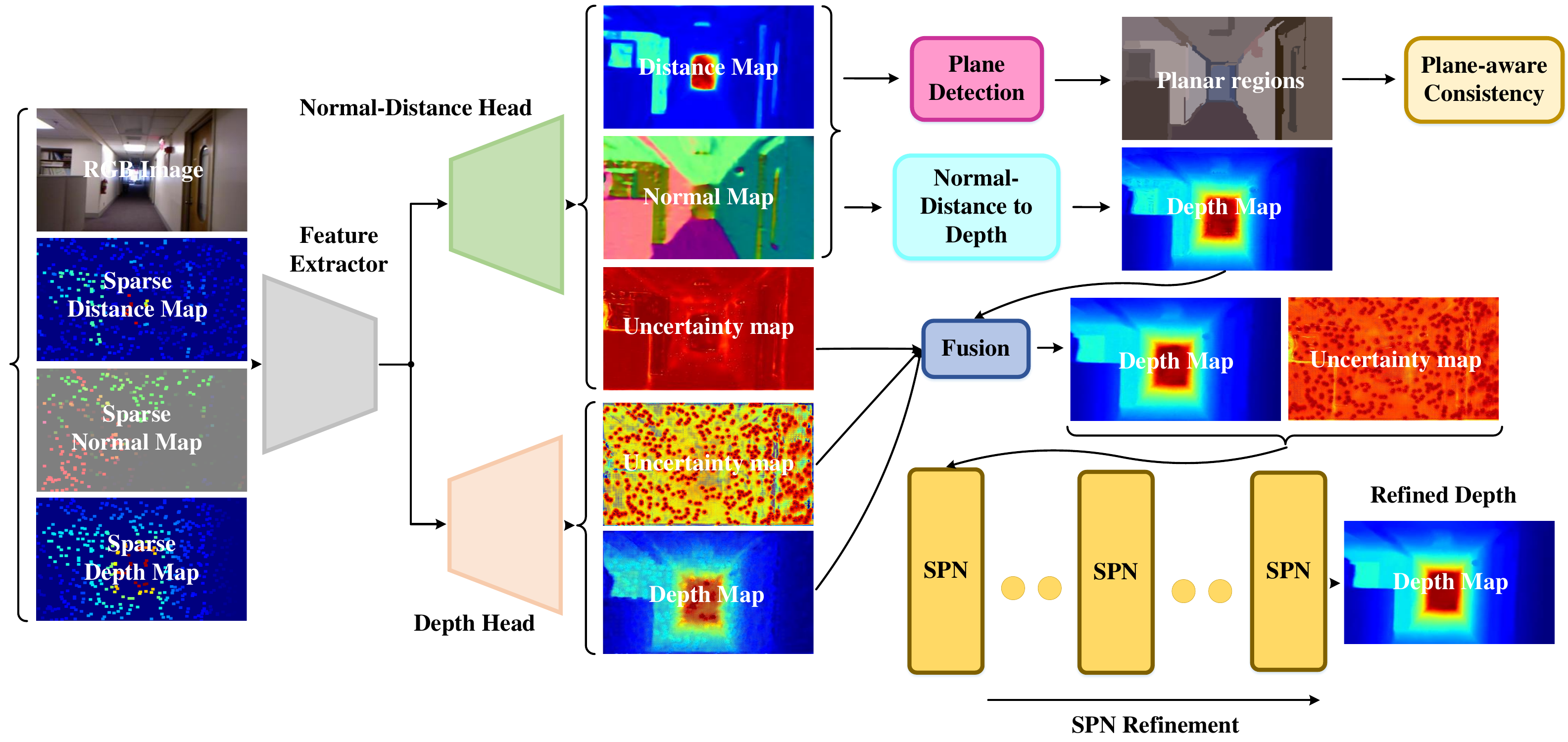}
	\caption{\textbf{Overview of the whole pipeline for depth completion}. Note that apart from the depth and uncertainty maps, the fused affinity map is also fed into the SPN to assist refinement. For simplicity, we have omitted this detail in the flowchart.  }
	\label{Fig12}
\end{figure*}

Utilizing this refinement module, the depth maps undergo an iterative enhancement until they ultimately reach convergence,
\begin{equation}\textbf{D}_1^{final}\left( \textbf{p} \right) \leftarrow \textbf{D}_1^t\left( \textbf{p} \right),\textbf{D}_2^{final}\left( \textbf{p} \right) \leftarrow \textbf{D}_2^t\left( \textbf{p} \right).
\end{equation}
Lastly, we upsample $\textbf{D}_1^{final}$ and $\textbf{D}_2^{final}$ to the full resolution, followed by an averaging operation to obtain the final output, 
\begin{equation}
	{\textbf{D}^{final}\left( \textbf{p} \right)} = 0.5(\textbf{D}_1^{final}\left( \textbf{p} \right) + \textbf{D}_2^{final}\left( \textbf{p} \right)).
\end{equation}
Notably, there is little difference in performance using simple averaging or uncertainty-based averaging for $\textbf{D}_1^{final}\left( \textbf{p} \right)$ and $\textbf{D}_2^{final}\left( \textbf{p} \right)$. This is reasonable since $\textbf{D}_1^{final}\left( \textbf{p} \right)$ and $\textbf{D}_2^{final}\left( \textbf{p} \right)$ are the final refined results according to the uncertainty maps. 

The training optimization objectives are detailed as follows:

\textbf{Depth loss}. We adopt a
scaled Scale-Invariant loss for depth supervision~\cite{lee2019big},
\begin{equation}
	\resizebox{0.85\hsize}{!}{${\mathcal{L}_\textbf{D}} = \sum\limits_{t = 1}^m {{\gamma ^{m - t}}} \left( \begin{array}{l}
			\kappa \sqrt {\frac{1}{{\left| {\bf{T}} \right|}}\sum\limits_{\bf{p}} {{{\left( {{{\bf{g}}_1^t}\left( {\bf{p}} \right)} \right)}^2} - \frac{\eta }{{{{\left| {\bf{T}} \right|}^2}}}{{\left( {\sum\limits_{\bf{p}} {{{\bf{g}}_1^t}\left( {\bf{p}} \right)} } \right)}^2}} }  + \\
			\kappa \sqrt {\frac{1}{{\left| {\bf{T}} \right|}}\sum\limits_{\bf{p}} {{{\left( {{{\bf{g}}_2^t}\left( {\bf{p}} \right)} \right)}^2} - \frac{\eta }{{{{\left| {\bf{T}} \right|}^2}}}{{\left( {\sum\limits_{\bf{p}} {{{\bf{g}}_2^t}\left( {\bf{p}} \right)} } \right)}^2}} } 
		\end{array} \right)$},
\end{equation}
where $\gamma$ is the decay factor and $m$ denotes the maximum iteration step, set to 0.85 and 3, respectively, ${\textbf{g}}\left( \textbf{p} \right) = \log {\textbf{D}}\left( \textbf{p} \right) - \log \textbf{D}^{gt}\left( \textbf{p} \right)$, \textbf{T} denotes a set of pixels containing valid values, $\kappa$ and $\eta$ are set to 10 and 0.85 following~\cite{lee2019big}.

\textbf{Normal loss}. We adopt a negative cosine loss to supervise surface normal~\cite{eigen2015predicting},
\begin{equation} {\mathcal{L}_\textbf{N}} = \frac{1}{{\left| {\bf{T}} \right|}}\sum\limits_\textbf{p} {1 - \textbf{N}\left( \textbf{p} \right) {\textbf{N}^{gt}}^{\rm{T}}\left( \textbf{p} \right)},  
\end{equation}

\textbf{Distance loss}. We adopt an L1 loss to supervise plane-to-origin distance,
\begin{equation}
	{\mathcal{L}_\mathcal{D}} = \frac{1}{{\left| {\bf{T}} \right|}}\sum\limits_\textbf{p} {\left| {\mathcal{D}\left( \textbf{p} \right) - {\mathcal{D}^{gt}}\left( \textbf{p} \right)} \right|}. \end{equation}

Due to the lack of surface normal and distance ground-truth in the NYU-Depth-v2 and KITTI datasets, we follow~\cite{qiu2019deeplidar} to acquire the ground-truth of surface normal  from depth ground-truth. Then, the distance ground-truth is obtained by ${\cal D}^{gt}\left( {\bf{p}} \right) = {\bf{D}}\left( {\bf{p}} \right)^{gt} {\bf{N}}\left( {\bf{p}} \right)^{gt}{{\bf{K}}^{ - 1}}{\widetilde {\textbf{p}}}$.

The \textbf{overall loss} is the combination of ${\mathcal{L}_\textbf{D}}$, ${\mathcal{L}_\textbf{N}}$, ${\mathcal{L}_\mathcal{D}}$, ${\mathcal{L}_\textbf{U}}$ and ${\mathcal{L}_{pc}}$,
\begin{equation}{\mathcal{L}_{overall = }}{\lambda _1}{\mathcal{L}_\textbf{D}} + {\lambda _2}{\mathcal{L}_\textbf{N}} + {\lambda _3}{\mathcal{L}_\mathcal{D}} + {\lambda _4}{\mathcal{L}_\textbf{U}} + {\lambda _5}{\mathcal{L}_{pc}},
\end{equation}
where ${\lambda _1}$, ${\lambda _2}$, ${\lambda _3}$, ${\lambda _4}$ and ${\lambda _5}$ are empirically set to 1, 5, 0.25, 1 and 0.01, respectively. 

\textbf{Network architecture.} We adopt Swin-L~\cite{Liu_2021_ICCV} as the feature extractor, unless specified otherwise. As for the KITTI official split with more training data, we choose SwinV2-L~\cite{liu2022swin} with a larger window size as the feature extractor to capture strong representations. The normal-distance head and the depth head share the same architecture except for the final prediction layer. These parts with the same architecture are designed according to NeWCRFs~\cite{Yuan_2022_CVPR}.  The normal-distance head predicts surface normal, distance and uncertainty maps, while the depth head predicts depth and uncertainty maps.

\subsection{Normal-Distance Assisted Depth Completion}

Given a sparse depth map $\textbf{D}^{sp}$ and an RGB image $\textbf{R}$, prior methods train a network to directly complete the sparse depth map into a dense one $\textbf{D}$ under the guidance of RGB image, 
\begin{equation} {N_\theta }: \textbf{D}^{sp}, \textbf{R} \to \textbf{D}. \end{equation} 
Unlike previous methods, our approach completes the sparse surface normal $\textbf{N}^{sp}$ and distance $\mathcal{D}^{sp}$, \begin{equation} {N_\theta }: \textbf{N}^{sp},\mathcal{D}^{sp}, \textbf{R} \to \textbf{N}, \mathcal{D}. \end{equation}
The surface normal and distance maps are regularized through the plane-aware consistency constraint and then converted into depth predictions using Eq.~\ref{eq4}. Similar to the estimation framework, we integrate a regular depth head based on CFormer~\cite{zhang2023completionformer} to improve the robustness of completion framework. 
Then, we fuse the depth maps from the normal-distance head and depth head based on the uncertainty maps. The depth uncertainty is modeled in the same way as the estimation framework. This fusion process is mathematically described as
\begin{equation}
	{\textbf{W}_1}\left( \textbf{p} \right) = \frac{{1 - {\textbf{U}_1}\left( \textbf{p} \right)}}{{2 - {\textbf{U}_1}\left( \textbf{p} \right) - {\textbf{U}_2}\left( \textbf{p} \right)}},
\end{equation}
\begin{equation}
	{\textbf{W}_2}\left( \textbf{p} \right) = \frac{{1 - {\textbf{U}_2}\left( \textbf{p} \right)}}{{2 - {\textbf{U}_1}\left( \textbf{p} \right) - {\textbf{U}_2}\left( \textbf{p} \right)}},
\end{equation}
\begin{equation}
	\textbf{D}(\textbf{p}) = {\textbf{W}_1}\left( \textbf{p} \right) {\textbf{D}_1}\left( \textbf{p} \right) + {\textbf{W}_2}\left( \textbf{p} \right) {\textbf{D}_2}\left( \textbf{p} \right),
\end{equation} 
\begin{equation}
	\textbf{U}(\textbf{p}) = {\textbf{W}_1}\left( \textbf{p} \right) {\textbf{U}_1}\left( \textbf{p} \right) + {\textbf{W}_2}\left( \textbf{p} \right) {\textbf{U}_2}\left( \textbf{p} \right).
\end{equation} 
 The fused depth and uncertainty maps are then input into the non-local spatial propagation network (NLSPN)~\cite{park2020non} for depth refinement.

\textbf{SPN refinement.} The spatial propagation of $	\textbf{D}^{t}(\textbf{p})$ at iteration $t$ using its non-local neighbors $\Omega(\textbf{p})$ is defined as
\begin{equation}
\resizebox{0.85\hsize}{!}{$
{\textbf{D}^{t + 1}}(\textbf{p}) = {\textbf{W}_\textbf{p}}\left( \textbf{p} \right){\textbf{D}^t}(\textbf{p}) + \sum\limits_{{\textbf{p}_{nei}} \in \Omega (\textbf{p})} {{\textbf{W}_{{\textbf{p}_{nei}}}}\left( \textbf{p} \right){\textbf{D}^t}({\textbf{p}_{nei}})}$},
\end{equation} 
where ${\textbf{W}_{{\textbf{p}_{nei}}}}(\textbf{p})$ denotes the affinity weight between the target pixel $\textbf{p}$ and its neighbor pixel ${\textbf{p}_{nei}}$, with value inside $\left( { - 1,1} \right)$, ${{\textbf{W}}_{\textbf{p}}}\left( {\textbf{p}} \right) = 1 - \sum\limits_{{\textbf{p}_{nei}} \in \Omega (\textbf{p})} {{\textbf{W}_{{\textbf{p}_{nei}}}}\left( \textbf{p} \right)}$ and signifies the degree to which the original depth will be conserved. Here, our normal-distance head and depth head also predicts the affinity maps, respectively, which are fused in the same fusion manner as the depth and uncertainty maps for SPN refinement. Additionally, the affinity map is modulated by the uncertainty map to reduce the negative impact of erroneous depth regions. Following $m$ iterations of spatial propagation, we acquire the final output $\textbf{D}^{final}(\textbf{p})$, where $m$ is set to 6 based on~\cite{zhang2023completionformer}. 

The training optimization objectives are detailed as follows:

\textbf{Depth loss.} We adopt a combined L1 and L2 loss for depth supervision~\cite{park2020non},
\begin{equation}
\resizebox{0.85\hsize}{!}{${\mathcal{L}_\textbf{D}} = \frac{1}{{\left| {\bf{T}} \right|}}\sum\limits_\textbf{p} {\left( {\left| {\textbf{D}^{final}\left( \textbf{p} \right) - {\textbf{D}^{gt}}\left( \textbf{p} \right)} \right| + {{\left| {\textbf{D}^{final}\left( \textbf{p} \right) - {\textbf{D}^{gt}}\left( \textbf{p} \right)} \right|}^2}} \right)}$}.
\end{equation} We note that only the final output is supervised following~\cite{park2020non,zhang2023completionformer}.

The normal loss, distance loss, uncertainty loss and plane-aware consistency loss are same to the estimation framework.

The \textbf{overall loss} is thus summarized as
\begin{equation}{\mathcal{L}_{overall = }}{\lambda _1}{\mathcal{L}_\textbf{D}} + {\lambda _2}{\mathcal{L}_\textbf{N}} + {\lambda _3}{\mathcal{L}_\mathcal{D}} + {\lambda _4}{\mathcal{L}_\textbf{U}} + {\lambda _5}{\mathcal{L}_{pc}},
\end{equation}
where ${\lambda _1}$, ${\lambda _2}$, ${\lambda _3}$, ${\lambda _4}$ and ${\lambda _5}$ are empirically set to 1, 5, 0.25, 1 and 0.01, respectively. 

\textbf{Network architecture.} We slightly modify the encoder in CFormer~\cite{zhang2023completionformer} as the feature extractor, which receives sparse surface normal map, sparse distance map, sparse depth map and RGB image. Besides, the design of normal-distance head and depth head mainly follows the decoder of CFormer. At the prediction layer, the former outputs surface normal, distance, uncertainty and affinity maps, while the latter outputs depth, uncertainty and affinity maps.

\begin{figure*}[!htb]
	\centering
	\includegraphics[width=0.97\linewidth]{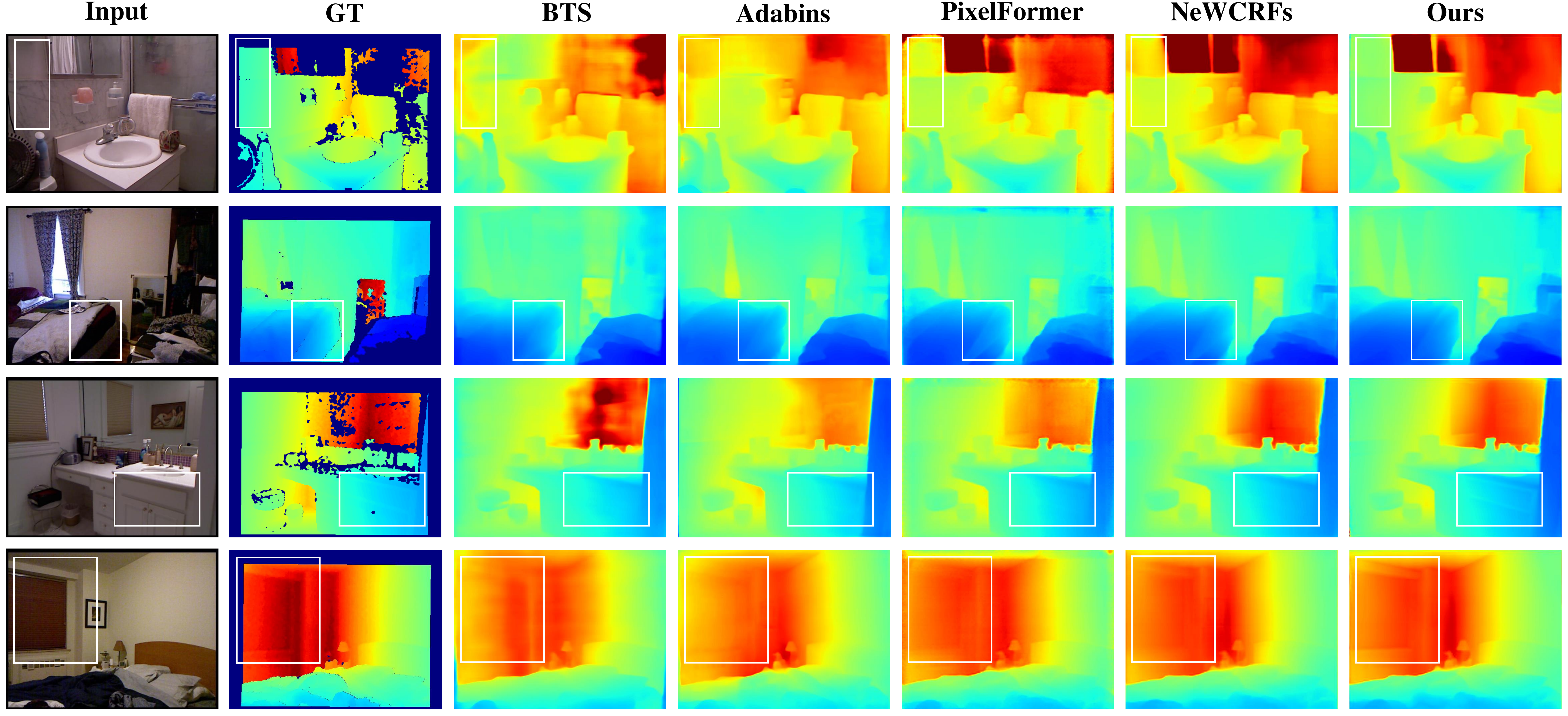}
	\caption{\textbf{Qualitative depth estimation results on the NYU-Depth-v2 dataset}. The white boxes show the regions to focus on. }
	\label{Fig6}
\end{figure*}

\section{Experiment}

\subsection{Datasets and Evaluation Metrics}
\textbf{NYU-Depth-v2 dataset} comprises data from indoor scenes, providing images and ground-truth depth maps with a resolution of $640 \times 480$ pixels. Regarding monocular depth estimation, we leverage 36253 images for training and 654 images for testing in line with~\cite{Yuan_2022_CVPR,agarwal2023attention}. The training and test resolution is $640 \times 480$. For depth completion, we employ 50000 images for training and 654 images for testing following~\cite{park2020non,zhang2023completionformer}. Moreover, the images and ground-truth depth maps are half-downsampled and center-cropped to a resolution of $304 \times 288$ during training and testing. The sparse depth, surface normal and distance maps are generated by randomly sampling from the corresponding ground-truths according to the sampling manner in~\cite{zhang2023completionformer}.

\textbf{KITTI dataset} consists of data from outdoor scenes, providing stereo images, sparse and ground-truth depth maps. The resolution is around $1241 \times 376$ pixels. As for monocular depth estimation, we employ two prevalent data splits, the Eigen split with 23488 training images and 697 test images~\cite{eigen2014depth}, as well as the official split with 85898 training images, 1000 validation images and 500 test images. The resolutions for training and testing are $1120 \times 352$ and $1216 \times 352$, respectively. Regarding depth completion, we use the official split with 85898 training images, 1000 validation images and 1000 test images. Due to the absence of LiDAR returns at the top of the depth map, the inputs undergo a bottom center-cropping operation for training based on~\cite{park2020non,zhang2023completionformer}, resulting in a resolution of $1216 \times 240$. The sparse surface normal and distance maps are calculated from sparse depth maps in the same way as ground-truth acquisition. The ground-truth depth maps for the official test set are not available and the results are generated by the online server.

\textbf{SUN RGB-D dataset} contains approximately 10K images, which are captured by four sensors in indoor scenes. We use this dataset to validate the generalization ability of monocular depth estimation in a challenging zero-shot setup, and evaluate pre-trained models on the official 5050 test images.

\begin{table*}[htb!]
	\begin{center}
		\renewcommand{\arraystretch}{1.3}
		\resizebox{1.87\columnwidth}{!}{\begin{tabular}{c || c || c c c c|| c c c}	
				\Xhline{1.2pt}
				Method &  Cap & Abs Rel $\downarrow$ & Sq Rel $\downarrow$ & RMSE $\downarrow$ & ${\textbf{\rm{log}}_{\bm{{10}}}}$ $\downarrow$ &  $\delta_{1.25}$ $\uparrow$ &  $\delta_{1.25^2}$ $\uparrow$& $\delta_{1.25^3}$ $\uparrow$ \\
				\hline						
				\hline
				Eigen~\textit{et al.}~\cite{eigen2014depth}& 0-10m& 0.158&-&0.641&-&0.769&0.950&0.988
				\\
				Fu~\textit{et al.}~\cite{fu2018deep}& 0-10m&0.115&-&0.509&0.051&0.828&0.965&0.992
				\\
				Qi et al.~\cite{qi2018geonet}&0-10m&0.128&-&0.569&0.057&0.834&0.960&0.990
				\\
				VNL~\cite{yin2019enforcing}& 0-10m&0.108&-&0.416&0.048&0.875&0.976&0.994
				\\
				BTS~\cite{lee2019big}& 0-10m&0.113&0.066&0.407&0.049&0.871&0.977&0.995
				\\
				Zhang~\textit{et al.}~\cite{zhang2020densely}& 0-10m&0.112&-&0.447&0.048&0.881&0.979&0.996
				\\
				DAV~\cite{huynh2020guiding}& 0-10m&0.108&-&0.412&-&0.882&0.980&0.996
				\\
				PWA~\cite{lee2021patch}& 0-10m&0.105&-&0.374&0.045&0.892&0.985&0.997
				\\
				Long~\textit{et al.}~\cite{Long_2021_ICCV}& 0-10m&0.101&-&0.377&0.044&0.890&0.982&0.996
				\\
				TransDepth~\cite{yang2021transformer}& 0-10m&0.106&-&0.365&0.045&0.900&0.983&0.996
				\\
				DPT~\cite{ranftl2021vision}& 0-10m&0.110&-&0.367&0.045&0.904&0.988&\textbf{0.998}\\
				Adabins~\cite{bhat2021adabins}& 0-10m&0.103&-&0.364&0.044&0.903&0.984&0.997
				\\
				P3Depth~\cite{patil2022p3depth}& 0-10m&0.104&-&0.356&0.043&0.898&0.981&0.996
				\\ 	
				Localbins~\cite{bhat2022localbins}&0-10m&0.099&-&0.357&0.042&0.907&0.987&\textbf{0.998}
				\\ 	
				DepthFormer~\cite{li2022depthformer}& 0-10m&0.096&-&0.339&0.041&0.921&0.989&\textbf{0.998}
				\\ 	
				NeWCRFs~\cite{Yuan_2022_CVPR}& 0-10m&0.095&0.045&0.334&0.041&0.922&\textbf{0.992}&\textbf{0.998}
				\\ 	
				PixelFormer~\cite{agarwal2023attention}& 0-10m&0.090&-&0.322&0.039&0.929&0.991&\textbf{0.998}
				\\ 					
				\hline
				\textbf{Ours} & 0-10m&\textbf{0.087} &\textbf{0.041}&\textbf{0.311}&\textbf{0.038}&\textbf{0.936}&0.991&\textbf{0.998}\\
				\Xhline{1.2pt}
		\end{tabular}}
	\end{center}
	\caption{\textbf{Quantitative depth estimation comparison on the NYU-Depth-v2 dataset}.  The best results are highlighted in \textbf{bold}.  }
	\label{table1}
\end{table*}

\textbf{Evaluation metrics.} Following~\cite{Yuan_2022_CVPR,agarwal2023attention}, we adopt metrics SILog, Abs Rel, Sq Rel, RMSE, iRMSE, RMSE log, ${\textbf{\rm{log}}_{\bm{{10}}}}$, $\delta_{1.25}$, $\delta_{1.25^{2}}$ and $\delta_{1.25^{3}}$ for monocular depth estimation. In line with~\cite{park2020non}, we adopt metrics REL, RMSE, iRMSE, MAE, iMAE, $\delta_{1.25}$, $\delta_{1.25^{2}}$ and $\delta_{1.25^{3}}$ for depth completion. Besides, we adopt metrics Mean, ${11.2^ \circ }$, ${22.5^ \circ }$ and ${30^ \circ }$~\cite{yu2020p} for surface normal comparison.

\subsection{Implementation Details}
We implement our frameworks in PyTorch~\cite{paszke2017automatic} on NVIDIA RTX A5000 GPUs. For monocular depth estimation, we adopt the Adam optimizer~\cite{kingma2015adam} where $\beta_1$ and $\beta_2$ are set to 0.9 and 0.999, respectively, along with a batch size of 8. The learning rate is scheduled using polynomial decay, starting from a base value of 2e-5 and decreasing to 2e-6. The total epochs are 25. For depth completion, we use the AdamW optimizer~\cite{loshchilov2017decoupled} with $\beta_1=0.9$ and $\beta_2=0.999$. The batch size and initial learning rate are set to 12 and 0.001, respectively. The model is trained for 72 epochs on the NYU-Depth-v2 dataset, with the learning rate being reduced by half at epochs 36, 48, and 60. As for the KITTI dataset, the model undergoes training for 100 epochs and we apply the learning rate decay by a factor of 0.5 at epochs 50, 60, 70, 80, and 90.  

\begin{figure}[!htb]
	\centering
	\includegraphics[width=1.0\linewidth]{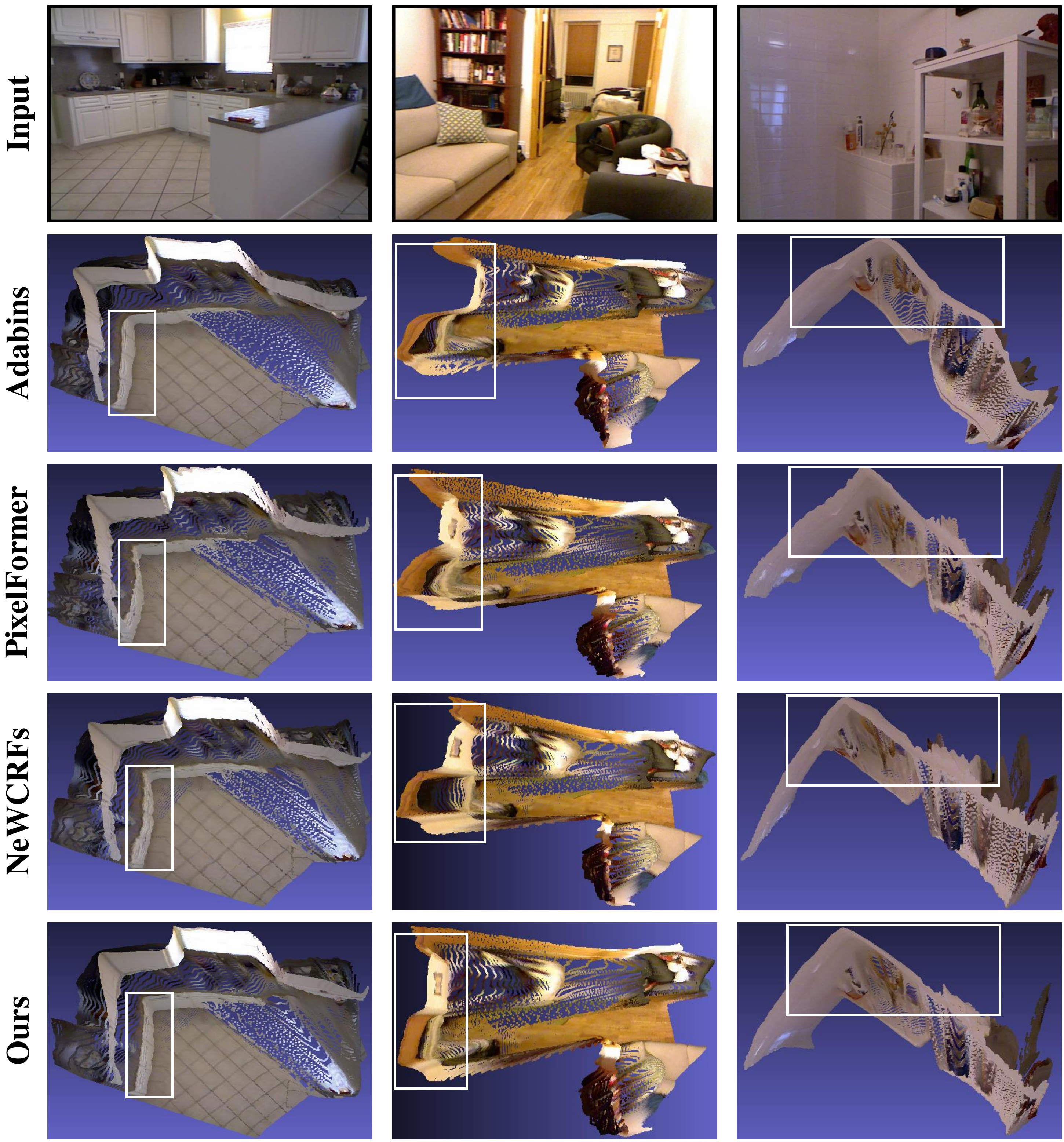}
	\caption{\textbf{Qualitative results of point cloud for monocular depth estimation on the NYU-Depth-v2 dataset}. The point clouds are captured from the top view.}
	\label{Fig8}
\end{figure}
\subsection{Monocular Depth Estimation}
\textbf{Comparison to previous competitors.} We first evaluate the proposed method on the NYU-Depth-v2 dataset. The results are summarized in Table~\ref{table1}. As we can see, our method exceeds previous competing methods on most metrics, with particularly notable results on Abs Rel and RMSE. Fig.~\ref{Fig6} provides qualitative depth comparison results, where our method excels in delineating planar regions while preserving local details,~\textit{e.g.}, boundaries. Besides, we present qualitative point cloud results in Fig.~\ref{Fig8}, which are obtained using depth maps. In particular, our point clouds preserve prominent geometric features such as planes and have fewer distortions compared to those generated by alternative methods. 

\begin{figure*}[!htb]
	\centering
	\includegraphics[width=0.97\linewidth]{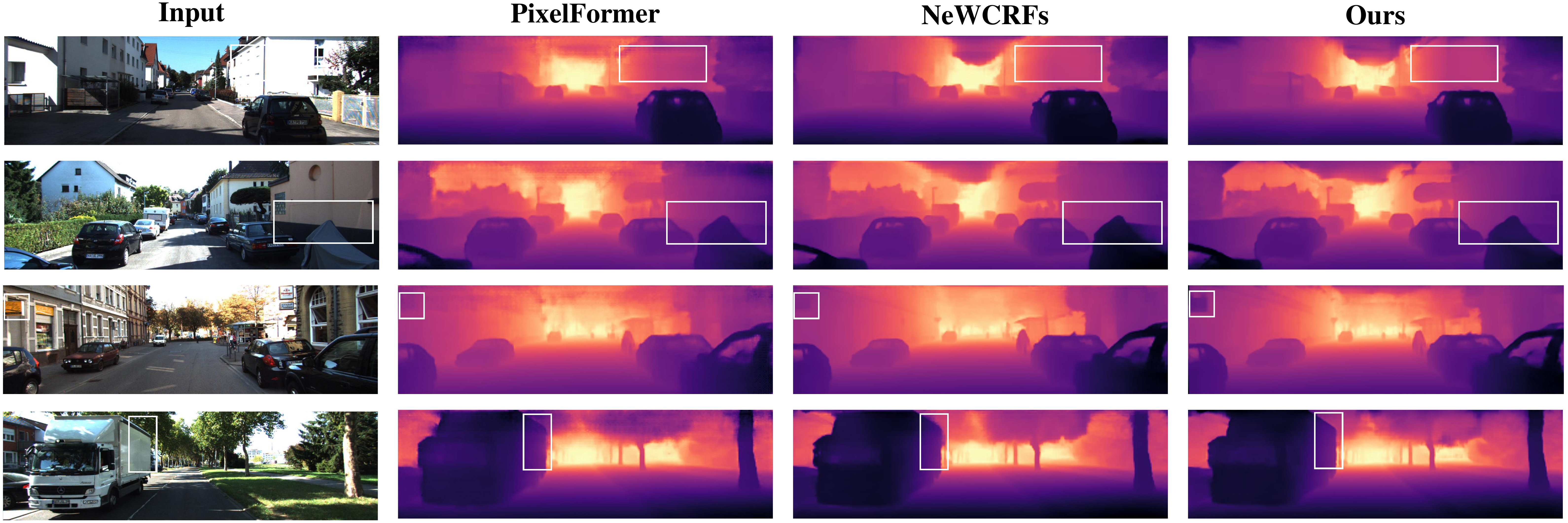}
	\caption{\textbf{Qualitative depth estimation results on the Eigen split of KITTI dataset}. }
	\label{Fig5}
\end{figure*}
\begin{figure}[!htb]
	\centering
	\includegraphics[width=1.0\linewidth]{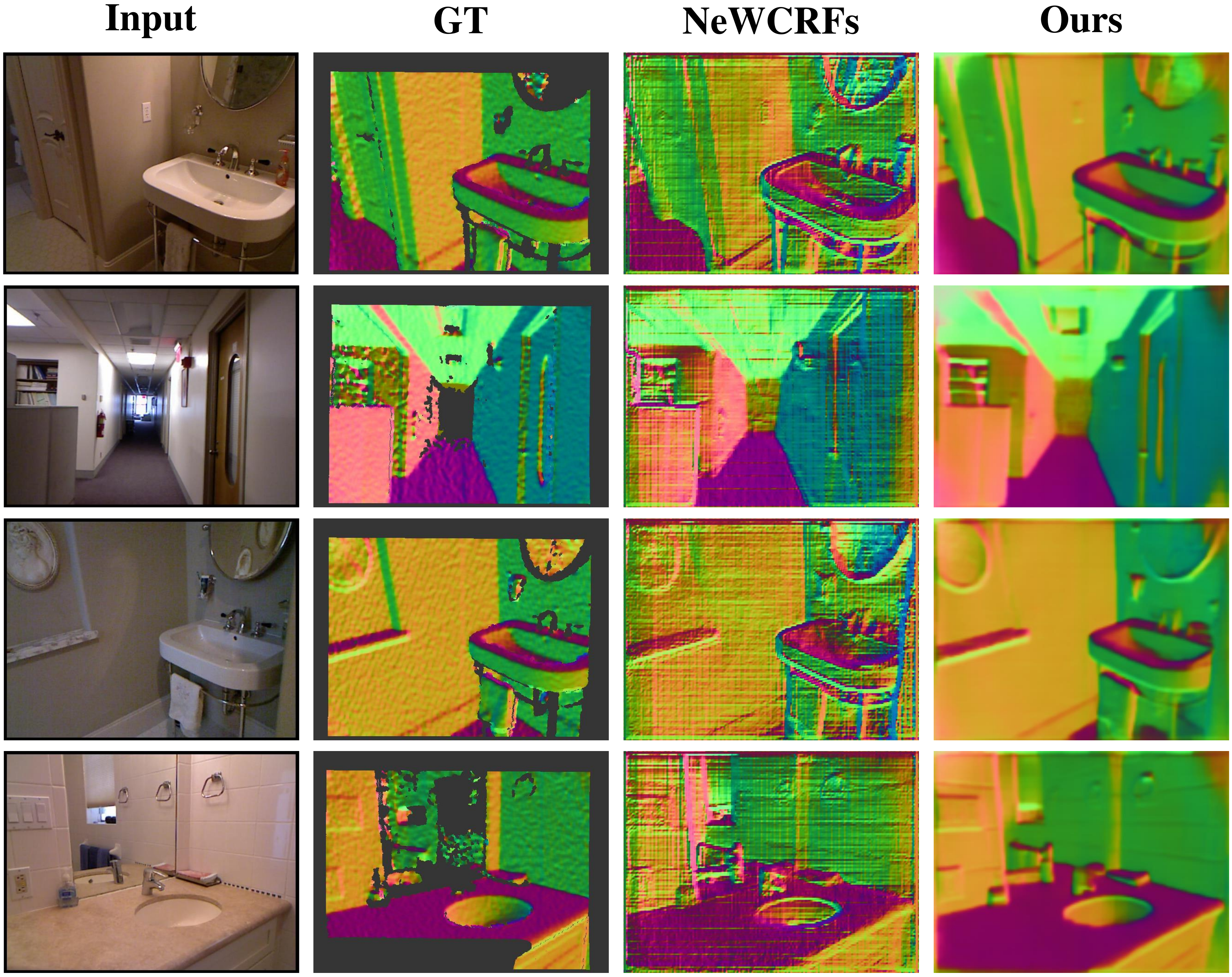}
	\caption{\textbf{Qualitative surface normal results on the NYU-Depth-v2 dataset}. }
	\label{Fig13}
\end{figure}

To further evaluate the structural properties of depth maps, we derive surface normal maps from depth maps leveraging a point-to-normal layer~\cite{yang2018unsupervised,yang2018lego}. The results are presented in Table~\ref{table12} and our method outperforms the compared methods in both cases, surface normal derived from depth and surface normal predicted by the normal-distance head. Fig.~\ref{Fig13} presents qualitative surface normal results. Although the ground-truth surface normal maps are slightly bumpy in the planar regions, affected by noise contained in the ground-truth depth maps, the predicted surface normal maps of our method exhibit good constant property in these planar regions. We attribute this to the imposed plane-aware consistency constraint.

We then evaluate the proposed method on the KITTI
dataset. The results on the Eigen split are presented in Table~\ref{table2} and our method outperforms previous leading methods by a large margin on most metrics, such as Sq Rel and RMSE, when the maximum depth is limited to 80m. For the 0-50m depth range, our method consistently outperforms the P3Depth. In addition, we observe that when compared to the PWA, P3Depth achieves superior performance within the 0-50m range but deteriorates significantly within the 0-80m range. \begin{table}[htb!]
	\begin{center}
		\renewcommand{\arraystretch}{1.3}
		\resizebox{0.92\columnwidth}{!}{\begin{tabular}{c|| c c c c   }	
				\Xhline{1.2pt}
				Method & Mean $\downarrow$ & ${11.2^ \circ }$ $\uparrow$ & ${22.5^ \circ }$ $\uparrow$ &  ${30^ \circ }$ $\uparrow$\\ 
				\hline						
				\hline
				BTS~\cite{lee2019big}&49.11&12.68&31.46&42.64\\
				Adabins~\cite{bhat2021adabins}&33.69&21.97&45.87&57.66\\
				NeWCRFs~\cite{Yuan_2022_CVPR}&33.95&20.43&45.19&57.55\\
				PixelFormer~\cite{agarwal2023attention}&37.77&16.89&39.07&51.07\\
				\hline
				\textbf{Ours} &30.91 &25.13&51.14&63.07\\
				\textbf{Ours}$\ddag$ &\textbf{24.41} &\textbf{31.82}&\textbf{61.66}&\textbf{72.97}\\
				
				\Xhline{1.2pt}		
		\end{tabular}}
	\end{center}
	\caption{\textbf{Quantitative surface normal comparison on the NYU-Depth-v2 dataset}. $\ddag$ indicates that the surface normal maps are predicted by the normal-distance head. } 
\label{table12}
\end{table}
\begin{table}[htb!]
\begin{center}
	\renewcommand{\arraystretch}{1.3}
	\resizebox{1.0\columnwidth}{!}{\begin{tabular}{c|| c c c c  }	
			\Xhline{1.2pt}
			Method & SILog $\downarrow$ & Sq Rel $\downarrow$ & Abs Rel $\downarrow$ & iRMSE$\downarrow$\\
			\hline						
			\hline
			PAP~\cite{zhang2019pattern}&13.08&10.27&2.72&13.95
			\\
			P3Depth~\cite{patil2022p3depth}&12.82&9.92&2.53&13.71
			\\
			VNL~\cite{yin2019enforcing}&12.65&10.15&2.46&13.02
			\\
			Fu~\textit{et al.}~\cite{fu2018deep}&11.77&8.78&2.23&12.98
			\\
			BTS~\cite{lee2019big}&11.67&9.04&2.21&12.23
			\\
			BA-Full~\cite{aich2021bidirectional}&11.61&9.38&2.29&12.23
			\\
			PackNet-SAN~\cite{guizilini2021sparse}&11.54&9.12&2.35&12.38
			\\
			PWA~\cite{lee2021patch}&11.45&9.05&2.30&12.32
			\\
			DepthFormer~\cite{li2022depthformer}&10.69&8.68&1.84&11.39
			\\
			NeWCRFs~\cite{Yuan_2022_CVPR}&10.39&8.37&1.83&11.03
			\\
			PixelFormer~\cite{agarwal2023attention}&10.28&8.16&1.82&10.84
			\\
			\hline
			\textbf{Ours} &\textbf{9.62} &\textbf{7.75}&\textbf{1.59}&\textbf{10.62}
			\\
			\Xhline{1.2pt}		
	\end{tabular}}
\end{center}
\caption{\textbf{Quantitative depth estimation comparison on the official split of KITTI dataset}. The SILog is the main ranking metric. Our method \textbf{ranks 1st} among all submissions on the KITTI depth prediction online benchmark at the submission time.} 
\label{table3}
\end{table}This discrepancy could be attributed to the prevalence of planar regions, such as roads, in nearby areas, while more distant regions are characterized by the high-curvature features such as bushes, trees, and other clutter, leading to the failure of the planarity prior. By contrast, our method relaxes such failure cases via the incorporation of a depth head, yielding promising results in both depth ranges. Fig.~\ref{Fig5} demonstrates qualitative depth results. As we can see, our method exhibits a higher proficiency in delineating planar regions,~\textit{e.g.}, wall and is less susceptible to color changes. 

\begin{figure*}[!htb]
\centering
\includegraphics[width=1.0\linewidth]{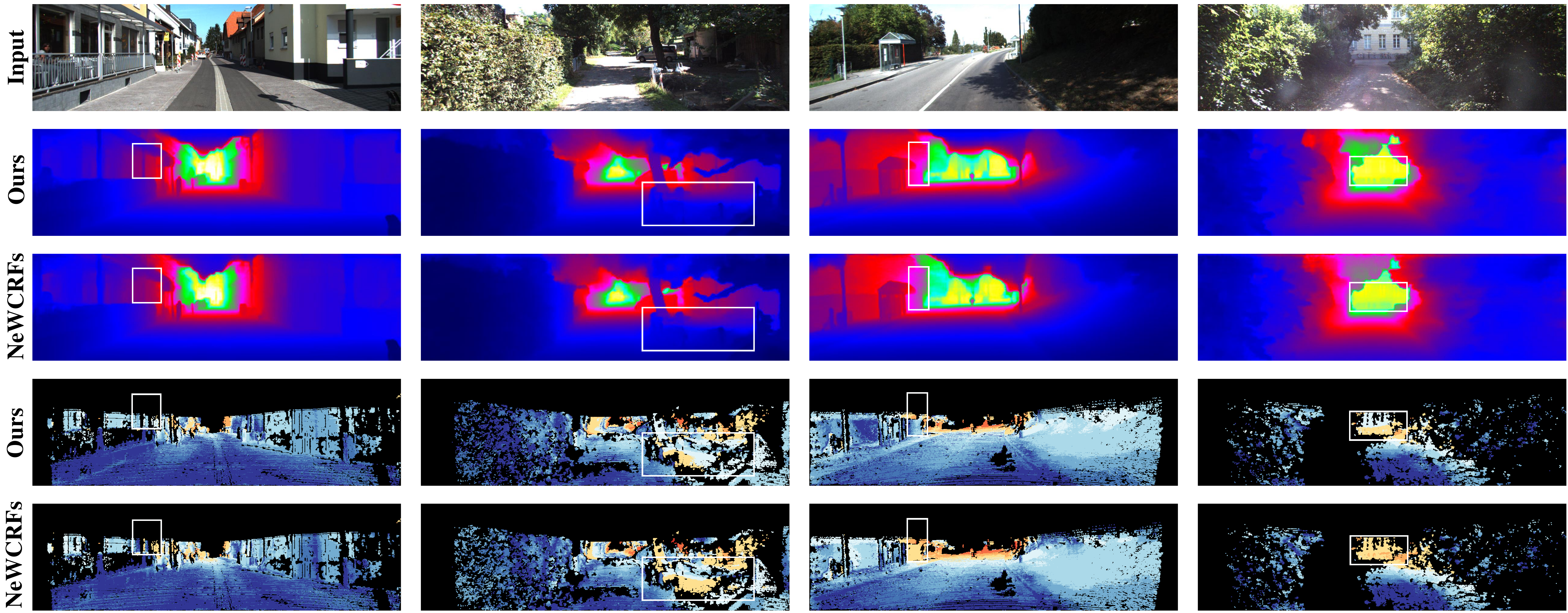}
\caption{\textbf{Qualitative depth estimation results on the official split of KITTI dataset}. The second and third rows correspond to the depth maps, while the fourth and fifth rows represent the error maps.}
\label{Fig3}
\end{figure*}
\begin{table*}[htb!]
\begin{center}
\renewcommand{\arraystretch}{1.3}
\resizebox{2.0\columnwidth}{!}{\begin{tabular}{c|| c || c c  c c || c c c }	
		\Xhline{1.2pt}
		Method & Cap & Abs Rel $\downarrow$ & Sq Rel $\downarrow$ & RMSE $\downarrow$ & RMSE log $\downarrow$ & $\delta_{1.25}$ $\uparrow$ & $\delta_{1.25^2}$ $\uparrow$& $\delta_{1.25^3}$ $\uparrow$ \\
		\hline						
		\hline
		Eigen~\textit{et al.}~\cite{eigen2015predicting}&0-80m&0.203&1.548&6.307&0.282&0.702&0.898&0.967
		\\
		Gan~\textit{et al.}~\cite{gan2018monocular}&0-80m&0.098&0.666&3.933&0.173&0.890&0.984&0.985
		\\
		Fu~\textit{et al.}~\cite{fu2018deep}&0-80m&0.072&0.307&2.727&0.120&0.932&0.984&0.994
		\\
		VNL~\cite{yin2019enforcing}&0-80m&0.072&-&3.258&0.117&0.938&0.990&0.998
		\\
		BTS~\cite{lee2019big}&0-80m&0.061&0.261&2.834&0.099&0.954&0.992&0.998
		\\
		Zhang~\textit{et al.}~\cite{zhang2020densely}&0-80m&0.064&0.265&3.084&0.106&0.952&0.993&0.998
		\\
		PWA~\cite{lee2021patch}&0-80m&0.060&0.221&2.604&0.093&0.958&0.994&\textbf{0.999}
		\\
		PackNet-SAN~\cite{guizilini2021sparse}& 0-80m&0.062&-&2.888&-&0.955&-&-\\
		TransDepth~\cite{yang2021transformer}&0-80m&0.064&0.252&2.755&0.098&0.956&0.994&\textbf{0.999}
		\\
		DPT~\cite{ranftl2021vision}& 0-80m&0.060&-&2.573&0.092&0.959&0.995&0.996\\
		Adabins~\cite{bhat2021adabins}&0-80m&0.058&0.190&2.360&0.088&0.964&0.995&\textbf{0.999}
		\\ 		
		P3Depth~\cite{patil2022p3depth}&0-80m&0.071&0.270&2.842&0.103&0.953&0.993&0.998
		\\
		DepthFormer~\cite{li2022depthformer}&0-80m&0.052&0.158&2.143&0.079&0.975&0.997&\textbf{0.999}
		\\
		NeWCRFs~\cite{Yuan_2022_CVPR}&0-80m&0.052&0.155&2.129&0.079&0.974&0.997&\textbf{0.999}
		\\
		PixelFormer~\cite{agarwal2023attention}&0-80m&0.051&0.149&2.081&0.077&0.976&0.997&\textbf{0.999}
		\\				
		\hline
		\textbf{Ours} &0-80m &\textbf{0.050}&\textbf{0.141}&\textbf{2.025}&\textbf{0.075}&\textbf{0.978}&\textbf{0.998}&\textbf{0.999}\\
		\hline
		Fu~\textit{et al.}~\cite{fu2018deep}&0-50m&0.071&0.268&2.271&0.116&0.936&0.985&0.995
		\\
		BTS~\cite{lee2019big}&0-50m&0.058&0.183&1.995&0.090&0.962&0.994&0.999
		\\
		Zhang~\textit{et al.}~\cite{zhang2020densely}&0-50m&0.061&0.200&2.283&0.099&0.960&0.995&0.999
		\\
		PWA~\cite{lee2021patch}&0-50m&0.057&0.161&1.872&0.087&0.965&0.995&0.999
		\\
		P3Depth~\cite{patil2022p3depth}&0-50m&0.055&0.130&1.651&0.081&0.974&0.997&0.999
		\\
		\hline
		\textbf{Ours} &0-50m&\textbf{0.048}&\textbf{0.107}&\textbf{1.513}&\textbf{0.071}&\textbf{0.981}&\textbf{0.998} &\textbf{1.000}\\
		\Xhline{1.2pt}
		
\end{tabular}}
\end{center}
\caption{\textbf{Quantitative depth estimation comparison on the Eigen split of KITTI dataset}. ``-'' means not applicable.} 
\label{table2}
\end{table*}
\begin{table}[htb!]
	\begin{center}
		\renewcommand{\arraystretch}{1.3}
		\resizebox{1.0\columnwidth}{!}{\begin{tabular}{c|| c c c c   }	
				\Xhline{1.2pt}
				Method & Abs Rel $\downarrow$ & RMSE $\downarrow$ & ${\textbf{\rm{log}}_{\bm{{10}}}}$ $\downarrow$ &  $\delta_{1.25}$ $\uparrow$\\ 
				\hline						
				\hline
				Chen~\textit{et al.}~\cite{chen2019structure}&0.166&0.494&0.071&0.757\\
				VNL~\cite{yin2019enforcing}&0.183&0.541&0.082&0.696\\
				BTS~\cite{lee2019big}&0.172&0.515&0.075&0.740\\
				Adabins~\cite{bhat2021adabins}&0.159&0.476&0.068&0.771\\
				P3Depth~\cite{patil2022p3depth}&0.178&0.541&-&0.698\\
				Localbins~\cite{bhat2022localbins}&0.156&0.470&0.067&0.777\\
				NeWCRFs~\cite{Yuan_2022_CVPR}&0.151&0.424&0.064&0.798\\
				PixelFormer~\cite{agarwal2023attention}&0.144&0.441&0.062&0.802\\
				\hline
				\textbf{Ours} &\textbf{0.137} &\textbf{0.411}&\textbf{0.060}&\textbf{0.820}\\
				
				\Xhline{1.2pt}		
		\end{tabular}}
	\end{center}
	\caption{\textbf{Generalization comparison on the SUN RGB-D dataset}. } 
\label{table5}
\end{table}
\begin{table}[htb!]
\begin{center}
	\renewcommand{\arraystretch}{1.3}
	\resizebox{1.0\columnwidth}{!}{\begin{tabular}{ c|| c c c c c c }
			\Xhline{1.2pt}
			Setting & PC & CIR & Abs Rel $\downarrow$ &  RMSE $\downarrow$  & ${\textbf{\rm{log}}_{\bm{{10}}}}$ $\downarrow$ & $\delta_{1.25}$ $\uparrow$\\
			\hline
			\hline				
			D&& & 0.095 & 0.334 & 0.041 & 0.922\\
			N$\mathcal{D}$&&& 0.092 & 0.324 & 0.039 & 0.926\\	
			N$\mathcal{D}$&\cmark&& 0.089 & 0.318 & 0.039 & 0.929\\	
			D$\&$N$\mathcal{D}$&\cmark&& 0.088 & 0.315 & 0.038 & 0.931\\
			D$\&$N$\mathcal{D}$&\cmark&\cmark& \textbf{0.087} & \textbf{0.311} & \textbf{0.038} & \textbf{0.936}\\	
			\hline
			D (planar)&& & 0.094 & 0.316 & 0.040 & 0.925\\
			N$\mathcal{D}$ (planar)&\cmark&& 0.088 & 0.301 & 0.038 & 0.934\\	
			D (non-planar)&& & 0.102 & 0.390 & 0.043 & 0.909\\
			N$\mathcal{D}$ (non-planar)&\cmark&& 0.104 & 0.394 & 0.044 & 0.908\\	
			\Xhline{1.2pt}			
	\end{tabular}}
\end{center}
\caption{\textbf{Ablation study for monocular depth estimation}. D$\&$N$\mathcal{D}$: integrated depth and normal-distance heads; D: depth head;  N$\mathcal{D}$: normal-distance head; PC: plane-aware consistency constraint; CIR: contrastive iterative refinement module. }
\label{table6}
\end{table}
\begin{figure}[!htb]
	\centering
	\includegraphics[width=1.0\linewidth]{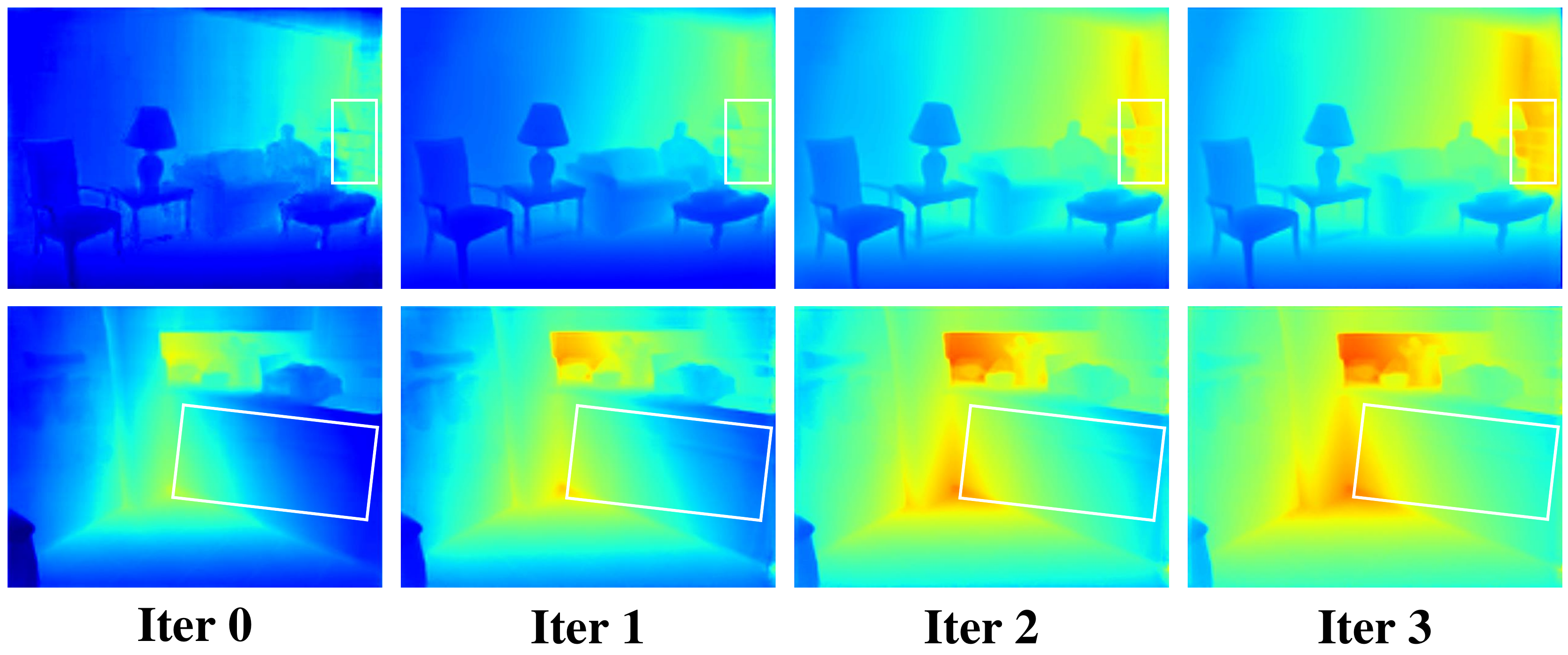}
	\caption{\textbf{Visualization of the refinement process in the contrastive iterative refinement module}. The first depth map in the upper row at iter 0 is from the normal-distance head, while the first depth map in the lower row at iter 0 is from the depth head. The white boxes indicate the regions to emphasize.}
	\label{Fig7}
\end{figure}

Table~\ref{table3} demonstrates the results on the official split. Once again, our method exceeds the compared methods, consistently showing improvements on all metrics. Additionally, the main ranking metric, SILog, experiences a considerable reduction, with the proposed method \textbf{ranking 1st} among all submissions on the KITTI depth prediction online benchmark at the time of submission. Fig.~\ref{Fig3} displays qualitative depth results, which are reported by the online server. The higher-quality depth maps from our method further emphasize our contributions.

\textbf{Zero-shot generalization.} Table~\ref{table5} demonstrates the comparison of generalization performance in a challenging zero-shot setting. The models are trained using the NYU-Depth-v2 dataset but evaluated using the SUN RGB-D dataset. The
impressive performance shows that the proposed physics-driven deep learning framework has strong generalization capabilities across different datasets. It indeed learns transferable features rather than merely memorizing training data statistics.

\begin{table}[htb!]
	\begin{center}
		\renewcommand{\arraystretch}{1.3}
		\resizebox{1.0\columnwidth}{!}{\begin{tabular}{ c|| c c c c c  }
				\Xhline{1.2pt}
				Method & RMSE & REL & $\delta_{1.25}$ & $\delta_{1.25^{2}}$ & $\delta_{1.25^{3}}$ \\ \hline \hline
				S2D~\cite{ma2018sparse} & 0.230 & 0.044 & 97.1 & 99.4 & 99.8 \\ 
				\cite{ma2018sparse}+Bilateral~\cite{barron2016fast} & 0.479 & 0.084 & 92.4 & 97.6 & 98.9 \\ 
				\cite{ma2018sparse}+SPN~\cite{liu2017learning} & 0.172 & 0.031 & 98.3 & 99.7 & 99.9 \\ 
				DepthCoeff~\cite{imran2019depth} & 0.118 & 0.013 & 99.4 & 99.9 & - \\ 
				CSPN~\cite{cheng2018depth} & 0.117 & 0.016 & 99.2 & 99.9 & \textbf{100.0} \\ 
				CSPN++~\cite{cheng2020cspn++} & 0.116 & - & - & - & - \\ 
				DeepLiDAR~\cite{qiu2019deeplidar} & 0.115 & 0.022 & 99.3 & 99.9 & \textbf{100.0} \\ 
				DepthNormal~\cite{xu2019depth} & 0.112 & 0.018 & 99.5 & 99.9 & \textbf{100.0} \\ 
				NLSPN~\cite{park2020non} & 0.092 & 0.012 & 99.6 & 99.9 & \textbf{100.0} \\
				ACMNet~\cite{zhao2021adaptive} & 0.105 & 0.015 & 99.4 & 99.9 & \textbf{100.0} \\ 
				TWISE~\cite{imran2021depth} & 0.097 & 0.013 & 99.6 & 99.9 & \textbf{100.0} \\
				RigNet~\cite{yan2022rignet} & 0.090 & 0.013 & 99.6 & 99.9 & \textbf{100.0} \\
				DySPN~\cite{lin2022dynamic} & 0.090 & 0.012 & 99.6 & 99.9 & \textbf{100.0} \\
				CFormer~\cite{zhang2023completionformer} & 0.090 & 0.012 & 99.6 & 99.9 & \textbf{100.0}  \\
				\hline
				\textbf{Ours} & \textbf{0.081} & \textbf{0.010} & \textbf{99.7} & \textbf{100.0} & \textbf{100.0} \\
				\Xhline{1.2pt}				
		\end{tabular}}
	\end{center}
	\caption{\textbf{Quantitative depth completion comparison on the NYU-Depth-v2 dataset}. We note that S2D~\cite{ma2018sparse} utilizes 200 sampled depth points per image as input, whereas the other methods use 500.}
	\label{table10}
\end{table}

\begin{table}[htb!]
	\begin{center}
		\renewcommand{\arraystretch}{1.3}
		\resizebox{1.0\columnwidth}{!}{\begin{tabular}{ c||c c c c  }
				\Xhline{1.2pt}
				\multirow{2}{*}{Method}
				&RMSE$\downarrow$ & MAE$\downarrow$ &iRMSE$\downarrow$ & iMAE$\downarrow$ \\
				 & (mm) & (mm) & (1/km) & (1/km)  \\ \hline \hline 
				CSPN~\cite{cheng2018depth} & 1019.64 & 279.46 & 2.93 & 1.15 \\ 
				S2D~\cite{ma2018sparse} & 814.73 & 249.95 & 2.80 & 1.21 \\ 
				DepthNormal~\cite{xu2019depth} & 777.05 & 235.17 & 2.42 & 1.13 \\ 
				DeepLiDAR~\cite{qiu2019deeplidar} & 758.38 & 226.50 & 2.56 & 1.15 \\ 
				FuseNet~\cite{chen2019learning} & 752.88 & 221.19 & 2.34 & 1.14 \\ 
				CSPN++~\cite{cheng2020cspn++} & 743.69 & 209.28 & 2.07 & 0.90 \\ 
				NLSPN~\cite{park2020non} & 741.68 & 199.59 & 1.99 & 0.84 \\
				ACMNet~\cite{zhao2021adaptive} & 744.91 & 206.09 & 2.08 & 0.90 \\
				TWISE~\cite{imran2021depth} & 840.20 & 195.58 & 2.08 & \textbf{0.82} \\
				RigNet~\cite{yan2022rignet} & 712.66 & 203.25 & 2.08 & 0.90 \\
				GuideFormer~\cite{rho2022guideformer} & 721.48 & 207.76 & 2.14 & 0.97 \\
				DySPN~\cite{lin2022dynamic} & 709.12 & \textbf{192.71} & \textbf{1.88} & \textbf{0.82} \\
				CFormer~\cite{zhang2023completionformer} & 708.87 & 203.45 & 2.01 & 0.88 \\
				\hline
				\textbf{Ours} & \textbf{698.71} & 192.75 & 1.89 & 0.83 \\
				\Xhline{1.2pt}				
		\end{tabular}}
	\end{center}
	\caption{\textbf{Quantitative depth completion comparison on the KITTI dataset}. The RMSE is the main ranking metric.
	}
	\label{table11}
\end{table}
\textbf{Ablation study.} To gain a deeper understanding of how each key component contributes to the performance, we conduct an ablation study and provide the results in Table~\ref{table6}. 

The results indicate that the normal-distance head surpasses the depth head when deployed independently, which appears to contradict the finding  in~\cite{patil2022p3depth} that predicting depth directly is superior to predicting plane coefficients. The reason behind may originate from our use of a more explicit normal-distance representation, as opposed to the implicit plane representation used in~\cite{patil2022p3depth}. This explicit representation enables us to introduce direct supervisory signals to guide the learning of surface normal and distance, hence resulting in more accurate depth estimates. Building upon the normal-distance head, we enforce the plane-aware consistency constraint on surface normal and distance maps, which consistently leads to improvements on most metrics. Then, we 
incorporate the depth head into our framework. The resulting improvements highlight the inherent complementarity of the depth estimates generated by these two heads. Finally, we append the contrastive iterative refinement module, thereby completing the entire framework and attaining the best results.

\begin{figure}[!htb]
	\centering
	\includegraphics[width=0.97\linewidth]{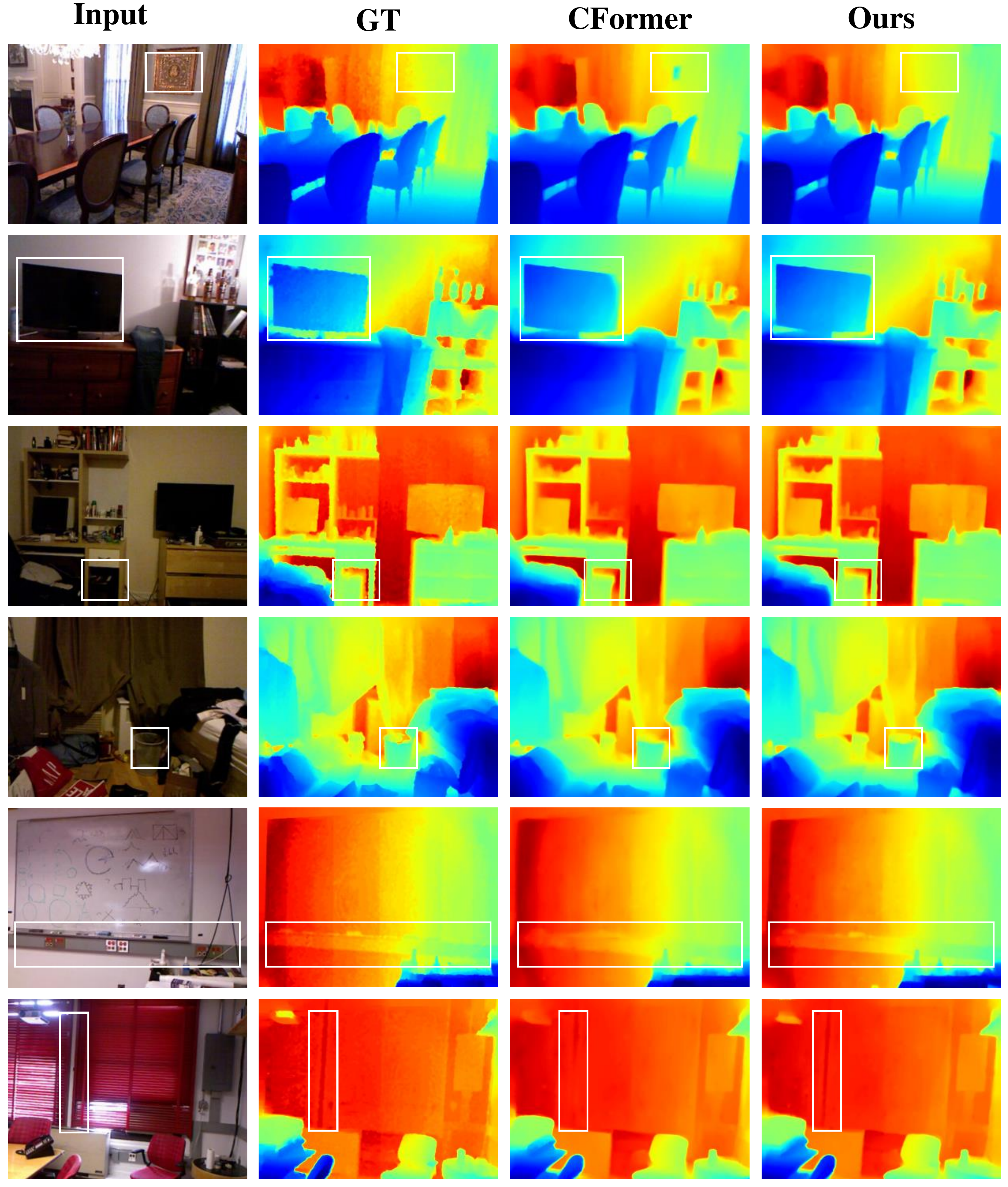}
	\caption{\textbf{Qualitative depth results for completion on the NYU-Depth-v2 dataset}. }
	\label{Fig10}
\end{figure}
\begin{figure}[!htb]
	\centering
	\includegraphics[width=1.0\linewidth]{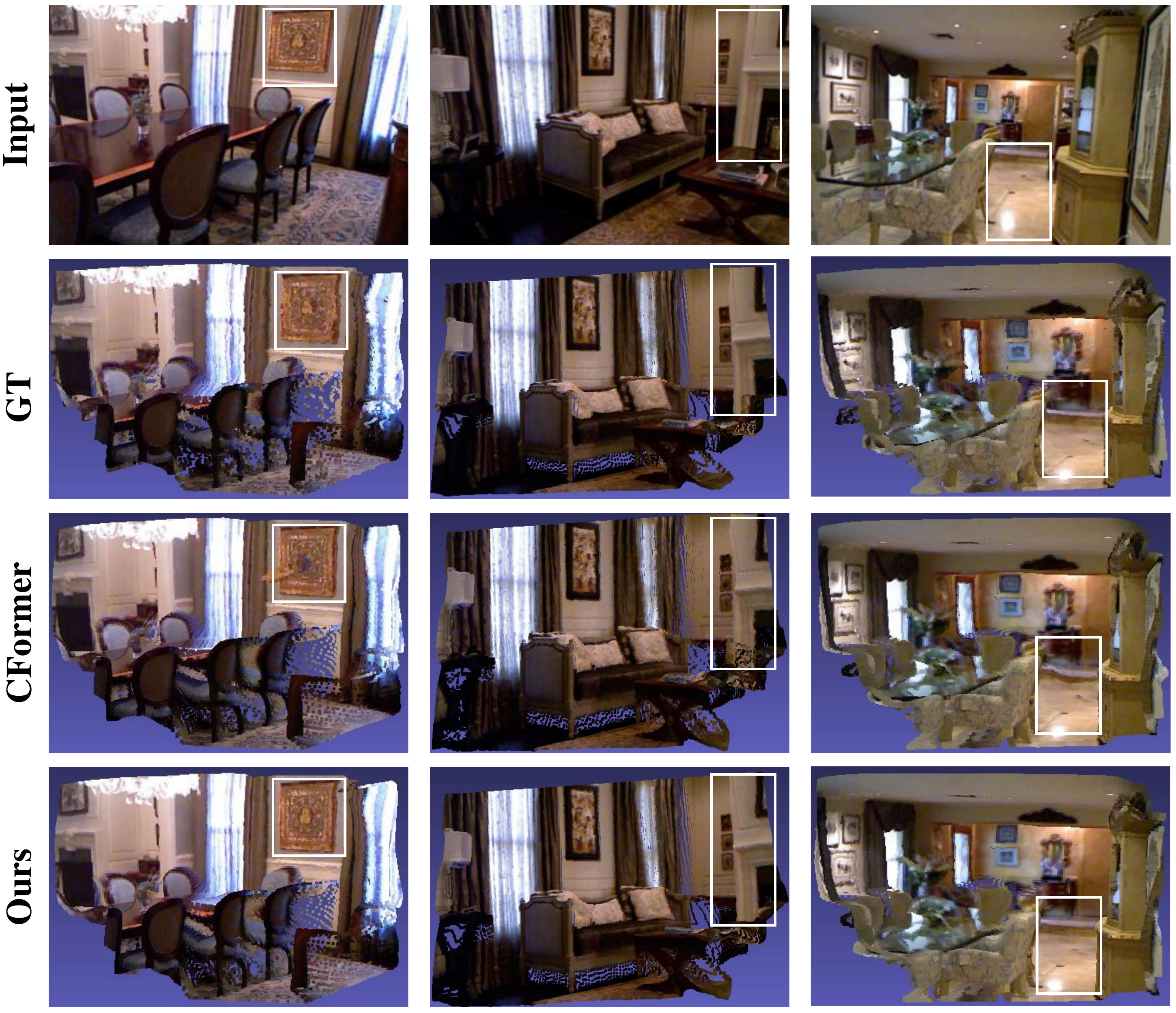}
	\caption{\textbf{Qualitative point cloud results for depth completion on the NYU-Depth-v2 dataset}. }
	\label{Fig14}
\end{figure}
\begin{table}[htb!]
	\renewcommand{\arraystretch}{1.3}
	\begin{center}
		\resizebox{1.0\columnwidth}{!}{
			\begin{tabular}{lc|ccccc}
				\Xhline{1.2pt}	
				& & \multicolumn{5}{c}{RMSE$\downarrow$} \\
				\multicolumn{2}{c|}{Method} & PackNet-SAN & GuideNet & NLSPN &CFormer & \textbf{Ours} \\
				\midrule
				\midrule
				& 0 & - & - & 0.562 & 0.490 & \textbf{0.471}  \\
				\cline{2-7}
				Sample & 50 & - & - & 0.223 & 0.208 & \textbf{0.187} \\
				\cline{2-7}
				Number & 200 & 0.155 & 0.142 & 0.129 & 0.127 & \textbf{0.118} \\
				\cline{2-7}
				& 500 & 0.120 & 0.101 & 0.092 & 0.090 & \textbf{0.081} \\
				\Xhline{1.2pt}	
			\end{tabular}
		}
	\end{center}
	\caption{\textbf{Sparsity study on the NYU-Depth-v2 dataset}. Evaluation with 0, 50, 200 and 500 samples. The compared methods are PackNet-SAN~\cite{guizilini2021sparse}, GuideNet~\cite{tang2020learning}, NLSPN~\cite{park2020non} and CFormer~\cite{zhang2023completionformer}.}
	\label{table8}
\end{table}
\begin{table}[htb!]
	\begin{center}
		\renewcommand{\arraystretch}{1.3}
		\resizebox{1.0\columnwidth}{!}{
			\begin{tabular}{c|c|cccc}
				\Xhline{1.2pt}	
				Scanning & \multirow{2}{*}{Method}
				&RMSE$\downarrow$ & MAE$\downarrow$ &iRMSE$\downarrow$ & iMAE$\downarrow$ \\
				Lines & & (mm) & (mm) & (1/km) & (1/km) \\
				\hline
				\hline
				\multirow{4}{*}{1} & 
				NLSPN & 3507.7 & 1849.1 & 13.8 & 8.9 \\
				& DySPN & 3625.5 & 1924.7 & 13.8 & 8.9 \\
				& CFormer & 3250.2 & 1582.6 & 10.4 & 6.6 \\
				& \textbf{Ours} & \textbf{3219.4} & \textbf{1552.9} & \textbf{10.3} & \textbf{6.5} \\
				
				\hline
				
				\multirow{4}{*}{4} & 
				NLSPN & 2293.1 & 831.3 & 7.0 & 3.4 \\
				& DySPN & 2285.8 & 834.3 & 6.3 & 3.2 \\
				& CFormer & 2150.0 & 740.1 & 5.4 & 2.6 \\
				& \textbf{Ours} & \textbf{2132.3} & \textbf{710.6} & \textbf{5.2} & \textbf{2.5} \\
				\hline
				
				\multirow{4}{*}{16} & 
				NLSPN & 1288.9 & 377.2 & 3.4 & 1.4 \\
				& DySPN & 1274.8 & 366.4 & 3.2 & 1.3 \\
				& CFormer & 1218.6 & 337.4 & 3.0 & \textbf{1.2} \\
				& \textbf{Ours} & \textbf{1172.8} & \textbf{322.1} & \textbf{2.9} & \textbf{1.2} \\
				\hline
				
				\multirow{4}{*}{64} & 
				NLSPN & 889.4 & 238.8 & 2.6 & 1.0 \\
				& DySPN & 878.5 & 228.6 & 2.5 & 1.0 \\
				& CFormer & 848.7 & 215.9 & 2.5 & \textbf{0.9}  \\
				& \textbf{Ours} & \textbf{819.6} & \textbf{209.1} & \textbf{2.3} & \textbf{0.9}  \\
				
				\Xhline{1.2pt}	
		\end{tabular}}
	\end{center}
	\caption{\textbf{Sparsity study on the KITTI dataset}. Evaluation with 1, 4, 16 and 64 scanning lines. The compared methods are NLSPN~\cite{park2020non}, DySPN~\cite{lin2022dynamic} and CFormer~\cite{zhang2023completionformer}.}
	\label{table7}
\end{table}
Fig.~\ref{Fig7} illustrates the refinement process in the contrastive iterative refinement module. As can be seen, when the number of iterations increases, the details of the locker in the depth maps from normal-distance head become increasingly distinct, while the kitchen cabinet surface in the depth maps from  depth head displays a more continuous appearance.

In addition, we compare the accuracy of depth maps from normal-distance
head and depth head in planar and non-planar regions, respectively, and the results support our standpoint.

\subsection{Depth Completion}

\textbf{Comparison to previous competitors.} Table~\ref{table10} presents the results on the NYU-Depth-v2 dataset. Despite the fact that the state-of-art performance has been nearly saturated for quite some time, such as from NLSPN to CFormer, our method is able to surpass the compared methods by a large margin. In particular, our method improves the RMSE by 10$\%$ and the REL by 16.7$\%$ over the CFormer. The large performance gap emphasizes that the proposed method significantly contributes to improving the accuracy. In Fig.~\ref{Fig10}, we provide qualitative depth results, and our method delivers more accurate estimates in planar regions and preserves depth edges. In Fig.~\ref{Fig14}, we demonstrate qualitative point cloud results. As can be seen, our method recovers the 3D structure reasonably. \begin{table}[htb!]
	\begin{center}
		\renewcommand{\arraystretch}{1.3}
		\resizebox{1.0\columnwidth}{!}{\begin{tabular}{ c|| c c c c c c }
				\Xhline{1.2pt}
				Setting & SPN & PC & RMSE $\downarrow$ &  REL $\downarrow$  & $\delta_{1.25}$ $\uparrow$ & $\delta_{1.25^2}$ $\uparrow$\\
				\hline
				\hline				
				D&& & 0.098 & 0.016 & 99.5 & 99.9\\
				N$\mathcal{D}$&&& 0.086 & 0.013 & 99.6 & \textbf{100.0}\\	
				D&\cmark& & 0.090 & 0.012 & 99.6 & 99.9\\
				N$\mathcal{D}$&\cmark&& 0.084 & 0.011 & 99.6 & \textbf{100.0}\\	
				N$\mathcal{D}$&\cmark&\cmark& 0.082 & 0.011 & \textbf{99.7} & \textbf{100.0}\\	
				D$\&$N$\mathcal{D}$&\cmark&\cmark& \textbf{0.081} & \textbf{0.010} & \textbf{99.7} & \textbf{100.0}\\
				\Xhline{1.2pt}				
		\end{tabular}}
	\end{center}
	\caption{\textbf{Ablation study for depth completion}. SPN: SPN refinement. D$\&$N$\mathcal{D}$: integrated depth and normal-distance heads; D: depth head;  N$\mathcal{D}$: normal-distance head; PC: plane-aware consistency constraint.}
	\label{table9}
\end{table}It is worth mentioning that in some cases, for instance, the highlighted pattern on the wall by white boxes in the first column, its point cloud recovered by ground-truth depth map shows slight errors in the description of internal structure, while our approach is capable of alleviating such failure cases, indicating that our physics-driven deep learning framework has robustness to the data noise.

\begin{figure*}[!htb]
	\centering
	\includegraphics[width=0.97\linewidth]{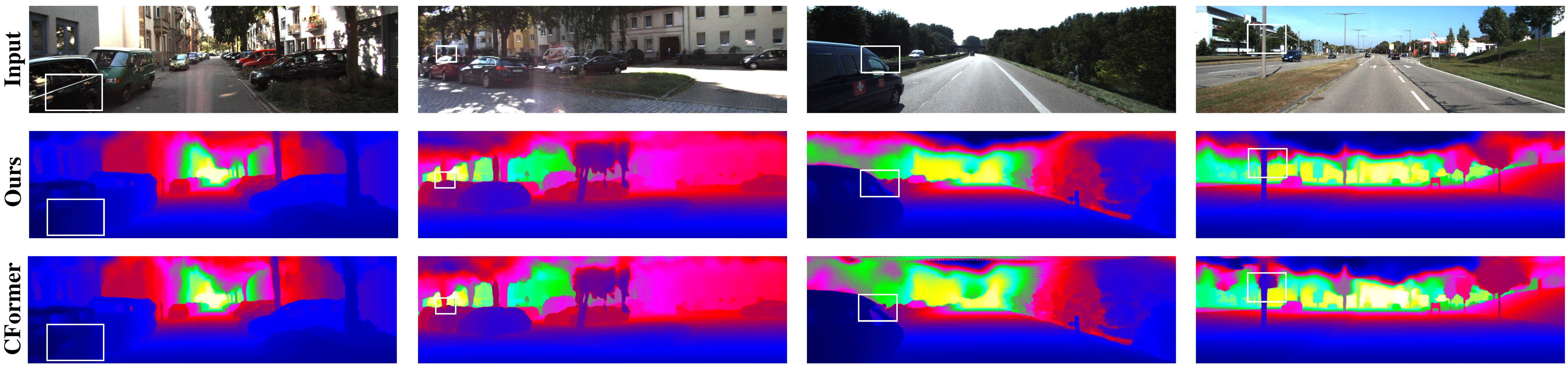}
	\caption{\textbf{Qualitative depth completion results on the KITTI dataset}.}
	\label{Fig9}
\end{figure*}

Table~\ref{table11} presents the results on the KITTI dataset. It can be seen that our method outperforms the compared methods on the main ranking metric RMSE. In Fig.~\ref{Fig9}, we demonstrate qualitative results from the online server and our method can achieve higher-quality depth maps than the CFormer.

\textbf{Sparsity study.} To showcase the robustness of our method when subjected to varying levels of data sparsity, we manually generate sparse data using different settings for training and testing. Regarding the NYU-Depth-v2 dataset, we randomly sample 0, 50, 200, and 500 points from the ground-truth depth, surface normal and distance maps to simulate various levels of sparsity. For the KITTI dataset, we follow~\cite{imran2021depth} and subsample the raw LiDAR scans in the azimuth-elevation space, reducing them into 1, 4 and 16 lines. Besides, we adopt the training set with 10000 images provided by~\cite{zhang2023completionformer} and evaluate on the KITTI validation set. The results of other approaches are from~\cite{zhang2023completionformer}. As summarized in Tables~\ref{table8} and~\ref{table7}, our method consistently surpasses the compared methods in all cases on both datasets.

\textbf{Ablation study.} In Table~\ref{table9}, we list the results of ablation study to better understand the impact of each key component. As can be seen, the SPN refinement is beneficial to improve performance whether using a normal-distance head or a depth head. In addition, the standalone performance of the normal-distance head surpasses that of the depth head by a large margin. However, this is not surprising considering that it could be easier to complete sparse surface normal and distance maps than it is to complete sparse depth maps. Specifically, the plane coefficients are piece-wise constant while the depth values tend to change from pixel to pixel. Once the sparse surface normal map or distance map has a value on a certain plane, one can easily complete this planar region by simply copying the value. Based on the normal-distance head, we apply the plane-aware consistency constraint for surface normal and distance maps, achieving improvements on most metrics. Finally, we combine the normal-distance head and depth head, and obtain the best results.

\section{Conclusion}
In this paper, we introduce new physics-driven deep learning frameworks for monocular depth estimation and completion based on the planar constraints in 3D scenes. The frameworks involve a normal-distance head to predict the piece-wise planar depth and a regular depth head to strengthen the robustness. In order to regularize the surface normal and distance maps, we develop a plane-aware consistency constraint. In addition, we develop a contrastive iterative refinement module to refine the depth maps. Extensive experiments indicate that the proposed method outperforms previous state-of-the-art monocular depth estimation and completion competitors on the NYU-Depth-v2, KITTI and SUN RGB-D datasets.

{\small
	\bibliographystyle{unsrt}
	\bibliography{egbib}
}

\end{document}